\documentclass[onecolumn]{svjourMy}          
\smartqed  

\usepackage{graphicx}
\usepackage{float}
\usepackage[cmex10]{amsmath}
\usepackage[comma,compress]{natbib}
\usepackage{caption}
\usepackage{epstopdf}
\usepackage{amssymb}
\usepackage{placeins}
\usepackage{multirow, makecell}
\usepackage{slashbox}
\usepackage{xcolor}
\usepackage[linesnumbered,ruled,boxed, vlined]{algorithm2e}
\usepackage{url}
\def\UrlBreaks{\do\*\do\-\do\~\do\'\do\"\do\-\do\=\do\r\do\/\do\m}

\begin{document}
\title{RelDenClu: A Relative Density based Biclustering Method for identifying non-linear feature relations}  

\subtitle{}

\titlerunning{Relative density-based biclustering}        

\author{Namita Jain         \and
		Susmita Ghosh		\and
		C A Murthy
}

\institute{Namita Jain \at
	Department of Computer Science and Engineering,
	Jadavpur University, Kolkata 700032, India
	\email{namita.saket@gmail.com}           
	\and
	Susmita Ghosh \at
	Department of Computer Science and Engineering,
	Jadavpur University, Kolkata 700032, India
	\email{susmitaghoshju@gmail.com}           
	\and
	C A Murthy \at
	Machine Intelligence Unit, Indian Statistical Institute, 203 Barrackpore Trunk Road,Kolkata 700108, India
}


\maketitle

\begin{abstract}

The existing biclustering algorithms for finding feature relation based biclusters often depend on assumptions like monotonicity or linearity. Though a few algorithms overcome this problem by using density-based methods, they tend to miss out many biclusters because they use global criteria for identifying dense regions. The proposed method, RelDenClu uses the local variations in marginal and joint densities for each pair of features to find the subset of observations, which forms the bases of the relation between them. It then finds the set of features connected by a common set of observations, resulting in a bicluster. 

To show the effectiveness of the proposed methodology, experimentation has been carried out on fifteen types of simulated datasets. Further, it has been applied to six real-life datasets. For three of these real-life datasets, the proposed method is used for unsupervised learning, while for other three real-life datasets it is used as an aid to supervised learning. For all the datasets the performance of the proposed method is compared with that of seven different state-of-the-art algorithms and the proposed algorithm is seen to produce better results. The efficacy of proposed algorithm is also seen by its use on COVID-19 dataset for identifying some features (genetic, demographics and others) that are likely to affect the spread of COVID-19. 

\keywords{Biclustering \and Relative density \and Marginal density \and Non-linear relationship}
\end{abstract}.
\section{Introduction}
Similarity within a bicluster can be interpreted in different ways like spatial closeness, the relation between features for selected observations. Most of the existing algorithms generally focus on spatial proximity between observations in the selected subspace. Relation-based biclustering methods mostly search for biclusters based on specific relations. While a few algorithms exist for finding non-linear relation-based biclusters, they search for biclusters with some constraint e.g., UniBic by \cite{unibic} finds monotonous relations and CBSC by \cite{CBSC2018} does not adjust for variations in marginal distributions, thereby limiting their ability to find various types of relation based biclusters. For example, UniBic will not be able to find a relation based periodic wave. CBSC finds more general relations, but if marginal densities have sparse and dense regions, the actual bicluster in the data may be not be obtained due to fragmentation. In this article, we have tried to overcome these problems.  
We propose a methodology for finding biclusters based on non-linear continuous feature relation. For each feature pair, we identify observations forming relationship between the two features by comparing joint density and marginal densities. Then, we check whether there exist more feature pairs connected with same set of observations forming related dense regions. Once this information is available, we have an adjacency matrix which allows us to find connected components. Thus, we will be able to find all the features connected with same set of observations. The resulting pair of sets (of observations and features) is called a bicluster. We have named this method RelDenClu, and it is seen to have several desirable properties like invariance to different permutations, scaling and translation. This allows results of RelDenClu to be consistent in spite of changes in data representation. The proposed technique of identifying related dense regions is the main contribution of this article.

We have discussed some of the existing biclustering methods in Section \ref{lits}. In Section \ref{novAlgo} we present the method of finding dense related regions using marginal and joint densities. Section \ref{detAlgo} presents entire algorithm in detail. Section \ref{methSec} describes datasets, method used for comparison and evaluation criterion. The results on simulated and real-life datasets are presented in Section \ref{resSec}. We have applied RelDenClu to analyse the effects of World Development Indicators on spread of COVID-19 in different regions of world, and presented the results in Section \ref{covidAn}, and conclude with Section \ref{Conc}.

\section{Related Work and Contributions}
\label{lits}
\subsection{Discussion on density based and relationship based biclustering methods}

\cite{HartiganDirect} pioneered biclustering by proposing a biclustering method based on constant value of elements. Subsequently, many biclustering algorithms were proposed with varying objectives and approaches. These include linear algebra based ones proposed by \cite{matfact2006} and \cite{Costeira1998}. Methods like spectral clustering use linear algebra to find biclusters in transformed space by \cite{Luxburg_spectral}. \cite{MITRA2006} used evolutionary computing, \cite{CTWCGetz} used iterative approach, while \cite{SAMBA2002} used graph-based approach. Detailed surveys on biclustering and subspace clustering have been done by \cite{KriegelBicluReview}, \cite{Prelic01052006}, \cite{ParsonsSubspaceReview}, \cite{surveyMaderia2004}.

In this article, we use a density-based approach for biclustering. Density-based biclustering algorithms such as SubClu by \cite{subclu} and CLIQUE by \cite{CLIQUE1998} find high-density regions in corresponding subspaces. These methods start by finding dense regions in one-dimensional space and grow biclusters using the apriori approach proposed by \cite{apriori1994}. These methods find biclusters based on spatial proximity while the proposed method finds biclusters based on feature relations. Further, the performance of these methods depends on the parameters used by the algorithm, while the proposed method, RelDenClu automatically finds out the parameters for density estimation for each feature pair. Thus, the user does not need to bother about underlying density estimation process and only needs to provide parameters directly related to biclustering.

As mentioned earlier, the main objective of RelDenClu is to find biclusters formed by features related to each other with a subset of observations as a base. Existing algorithms with this objective include the algorithm proposed by \cite{cheng2000biclustering}, multiplicative algorithm FABIA proposed by \cite{FABIA2010}. Here a row can be obtained by multiplying another with a constant. Unlike our proposed method, these methods assume that the relation between features is linear or multiplicative. 

More generalized biclustering methods based on an arbitrary relationship between features also exist in literature. These include algorithms like UniBic by \cite{unibic}, based on longest common subsequence. UniBic can find biclusters based on monotonous relationships between features. Curler by \cite{curler2005} finds such non-linear clusters by combining a density-based approach with the principal component method. The method finds high-density Gaussian regions using Expectation Maximization and combines them using the relation between directional information of resultant biclusters. OPSM proposed by \cite{OPSM2005} finds order-preserving submatrices. 

A more recent algorithm named CBSC proposed by \cite{CBSC2018} identifies non-linear relationships between pairs of dimensions using a density-based approach. However, CBSC tends to produce fragmented biclusters when marginal densities are highly variable. 

The proposed algorithm, RelDenClu, overcomes the problem of fragmentation by adapting to local variations in density along each dimension. It finds non-linear continuous relationship based biclusters and does not make assumptions of monotonicity, linearity or fixed relationship between features.

\subsection{Contributions}
The proposed method, RelDenClu finds biclusters based on continuous non-linear relationships using marginal and joint densities and, information derived from 2D spaces corresponding to each pair of features. The advantages of this approach are listed below. The first point in this list is discussed in Section \ref{choiceBinPrama} and the last three points have been elaborated using a toy example in Section \ref{toyex}.
\begin{enumerate}
	\item It does not expect the user to provide parameters for finding high-density regions.
	\item It does not assume that the relationship between features has a particular form like linear or multiplicative.
	\item It does not assume that the relationship between features is monotonous. 
	\item It avoids the fragmentation of biclusters by adapting for variations in marginal densities.
\end{enumerate}

\section{Proposed Biclustering Method: RelDenClu}
\label{novAlgo}
In this section, we define the objective of RelDenClu and provide relevant definitions. We also present the procedure for finding dense sets for each pair of features, which is the main contribution of the present article.
\subsection{Objective of the proposed biclustering method}
\label{objAlgo}
The bicluster obtained by RelDenClu is a submatrix in which columns are related directly or indirectly. Let the dataset be denoted by a matrix $D$ containing $N$ rows each corresponding to observation and $M$ columns each corresponding to a feature. In this article, a direct relationship between two columns of a matrix means the following: Dependence exists between features corresponding to these two columns when the subset of observations contained in the bicluster is used as the base. Two columns are connected indirectly if there exists a chain of columns connecting them, such that each consecutive pair of columns in this chain is directly connected. This has been stated in the definitions below.

\begin{definition}
	For a matrix $A$ let the $i^{th}$ column be denoted by $A_{*, i}$.
	In this article, we call the $i^{th}$ and the $j^{th}$ columns of a matrix to be directly connected if there is a dependence between these two columns. Perfect dependence occurs when either $A_{*,i}$ can be used to determine $A_{*,j}$, or vice-versa.
\end{definition}
Imagine each direct relationship between features to be an edge connecting two vertices, each representing a column. Then, two columns are indirectly connected if they are connected by a path. Note that the direct or indirect connection between the columns exists only on the base of rows contained in the submatrix.
\begin{definition}
	We call the $i^{th}$ and the $j^{th}$ columns of a matrix $A$ to be connected if these two columns are directly connected or there exist columns $A_{*,i_1}, \cdots, A_{*,i_k} $, such that 
	1) each column in this list is directly connected to the consecutive column in the list (e.g. $A_{*,i_2}$ is directly connected to $A_{*,i_3}$),
	2) $A_{*,i}$ is directly connected to $A_{*,i_1}$, 	and 3)   $A_{*,i_k}$ is directly connected to $A_{*,j}$.
\end{definition}

Using the definition of connected features we now define a bicluster as follows:
\begin{definition}
	A bicluster in the data matrix $D$ is a pair of two sets given by $O=\{oi_1, oi_2, \cdots, oi_n\}$ and $F=\{fi_1, fi_2, \cdots, fi_m\}$. The matrix $D$ restricted to observations in $O$ and features in $F$, gives us a submatrix $A$ corresponding to the bicluster $<O, F>$. More specifically, the submatrix corresponding to $<O,F>$ is given by $A$ where the $q^{th}$ element of the $p^{th}$ row is given by, $A[p,q]=D[oi_p, fi_q]$. A bicluster is said to be relation-based if all the columns of the corresponding submatrix are connected, either directly or indirectly. 
\end{definition}
It may be noted that the proposed method finds biclusters based on the relation between feature pairs, and so it can result in biclusters with disconnected regions. Thus, the proposed method can be seen as grouping related features based on observations. 

To find biclusters based on definitions given above, we need to check whether two features are related to each other for any set of observations. A method to find related subsets of observations for a given feature pair is presented in the following section.
\subsection{Proposed technique of finding a set of observations having dependence for a given feature pair}
\label{proposedHighDense}

This section describes a novel method for identifying subsets of observations which show dependence between two features, by comparing joint distribution and marginal distributions using histogram technique.

The data in the Euclidean space is normalized to space $[0,1]\times[0,1]$. Either a grid or rolling window is used to calculate the density. Data is analyzed by considering small rectangular regions called cells. The terminology used in this section is as follows: The region given by $(v_x - xlen/2, v_x + xlen/2]\times(v_y - ylen/2, v_y+ylen/2]$ is said to be a cell of size $(xlen, ylen)$ centred at point$(v_x, v_y)$. The phrase ``corresponding marginal cells'' refers to rectangular regions: $(v_x - xlen/2,   v_x + xlen/2]\times[0,1]$ and $[0,1] \times (v_y - ylen/2, v_y+ylen/2]$. Here we use the term `density' for `histogram density estimate' i.e., density is given by the number of observations in a cell divided by product of cell area and total number of observations. A cell is said to be dense if its density is higher than the average density of the corresponding marginal cells. 

As examining cells centered at each observation may not be computationally feasible for large datasets, we use slightly different schemes for small and large datasets. This allows us to achieve a balance between accuracy and ease of computation. Whether we consider a dataset to be large or small depends on computational power.

\subsubsection{Finding related dense regions for small datasets}
\label{smallHighDense}
For small datasets, cell centered around each observation is examined. Note that, we will have many overlapping cells. For each observation with coordinates $(v_x, v_y)$ we find observations lying in the cell centred on it. More precisely, we check the cell given by $(v_x-xlen/2, v_x + xlen/2]\times(v_y-ylen/2, v_y+ylen/2]$ to see if its density is greater than the average density of both the marginal cells, which are the cells given by $(v_x-xlen/2, v_x + xlen/2]\times(0, 1]$ and  $(0,1]\times(v_y-ylen/2, v_y+ylen/2]$. If this condition is found to be true and the density of the cell is also higher than the average density of the entire region, we mark the cell centred at $(v_x, v_y)$ as `dense'. 
Two dense cells are merged only if the center of each cell lies within the other cell and the average density of the overlapping region between two cells is high. Density is considered high if it is not less than the density of each of the marginal cells corresponding to the two overlapping cells and also not less than the average density of the entire space.
Repeating these steps iteratively allows us to find larger connected dense regions. Algorithm \ref{algodense} provides pseudo-code for finding dense regions as described here.
	
The size of a cell along a dimension $X$ is calculated using maximal separation along $X$. For each pair of dimensions $X$ and $Y$ let the maximal separations be $s_x$ and $s_y$, respectively. The intervals used to find the dense regions along $X$ and $Y$ are given by $xlen={s_x}^c$ and $ylen={s_y}^c$, where $c<0.5$ and is close to $0.5$. We have taken $c=0.4999$. The reason behind choosing this value has been discussed in Section \ref{choiceBinPrama}. 

 As the maximal separation does not go to zero for datasets having non-continuous distribution, this method cannot be used for such datasets. Therefore, for such datasets and for large datasets we present a simpler method which does not use maximal separation. This is discussed in the following section.
\subsubsection{Finding related dense regions for large datasets}
\label{largeHighDense}
For larger datasets having non-continuous distribution, the method elaborated in Section \ref{smallHighDense} would not be feasible, as the number of observations in the neighbourhood of each observation has to be calculated. To overcome this problem we present a simplified method for identifying dense regions in large datasets. For large datasets, we simply take $3\log(N)$ equal intervals along each axis, where $N$ is the number of observations in the dataset. Since only disjoint cells are examined, execution time is reduced. We decide whether a cell is dense using the same criterion as small datasets i.e., the density of the cell should be greater than densities of marginal cells as well as the average density of the entire space. Two dense regions are merged if they are adjacent to each other, horizontally, vertically or diagonally. A dense region can have at most 8 dense regions connected to it. After finding dense regions in 2D space, noise is removed by using the overlap between three two-dimensional spaces for every set of three features. Thus, redundancy introduced by using three features, is used to weed out the noise. A detailed description of this step can be found in Section \ref{mergebase}.

\subsubsection{The proposed method on a toy example}
\label{toyex}

\begin{figure}[!htbp]
	\begin{minipage}{\linewidth}
		
		\begin{minipage}[t]{0.45\linewidth}

			\includegraphics[width=\linewidth]{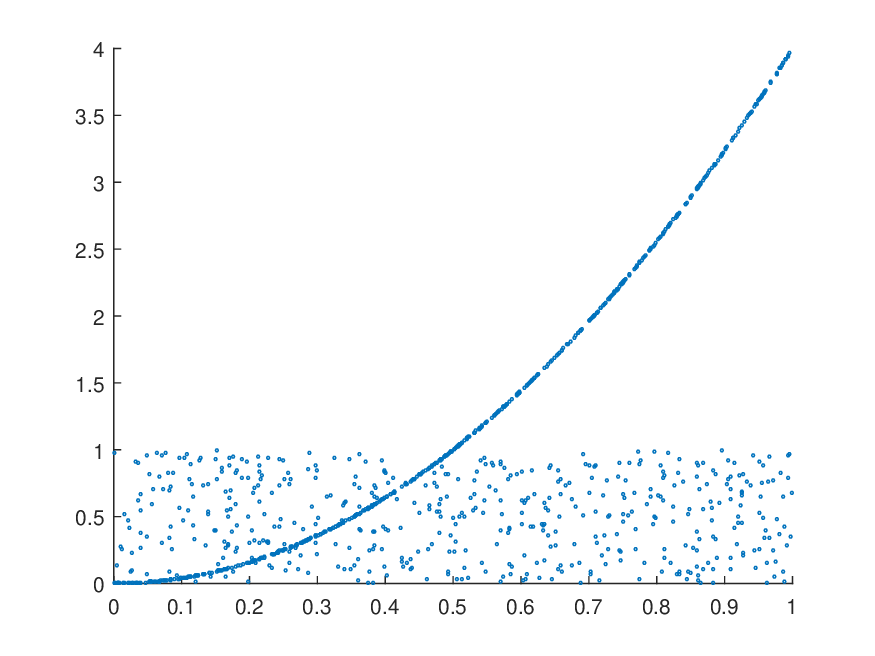}
			\label{nUdata}
			
		\end{minipage}
		\hspace{0.05\linewidth}
		\begin{minipage}[t]{0.45\linewidth}
			
			\includegraphics[width=\linewidth]{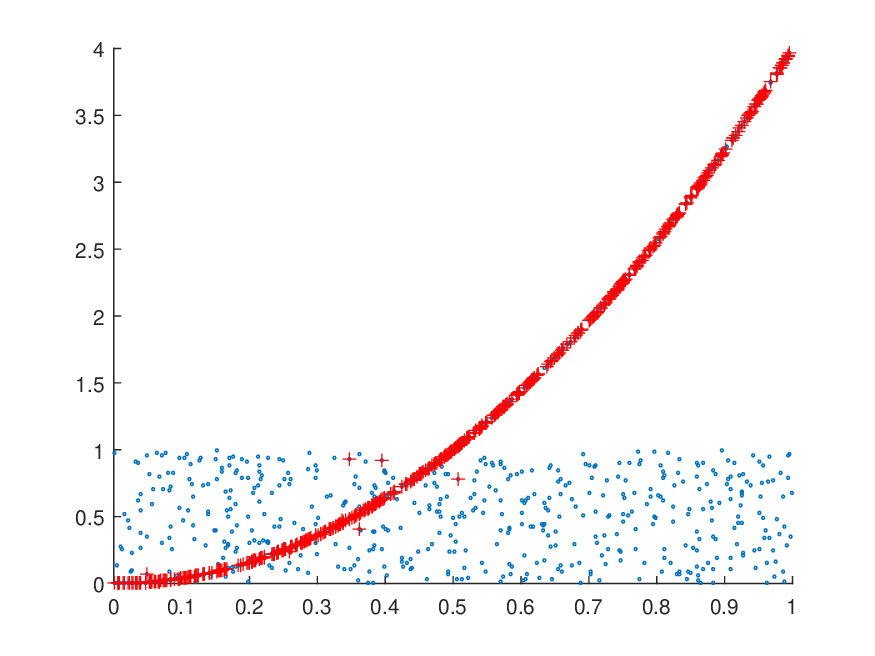}
			
			\label{nUdatares}
			
		\end{minipage}

		\begin{minipage}[t]{0.45\linewidth}
			
			\includegraphics[width=\linewidth]{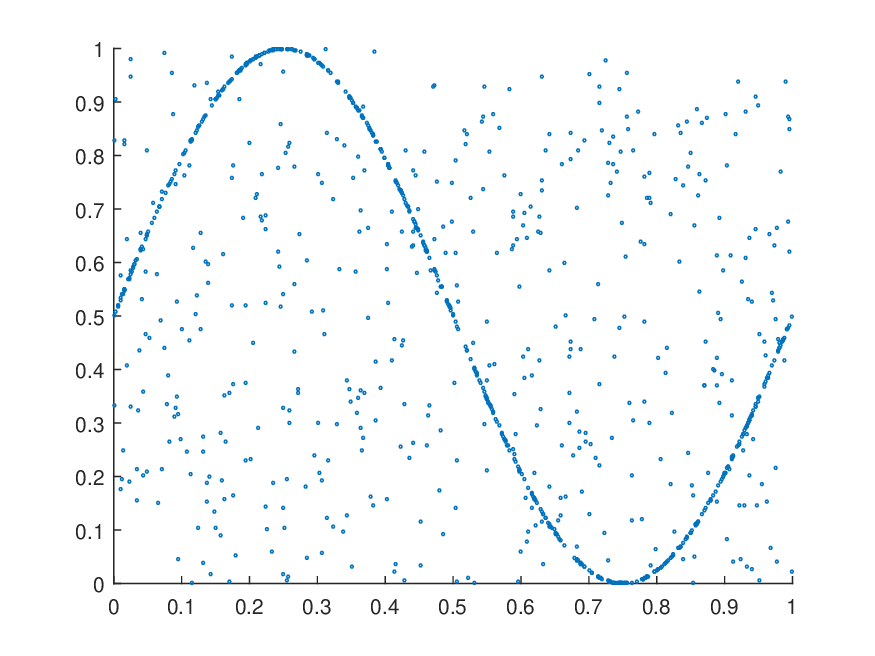}
			\label{datanonM}
			
		\end{minipage}
		\hspace{0.05\linewidth}
		\begin{minipage}[t]{0.45\linewidth}
			
			\includegraphics[width=\linewidth]{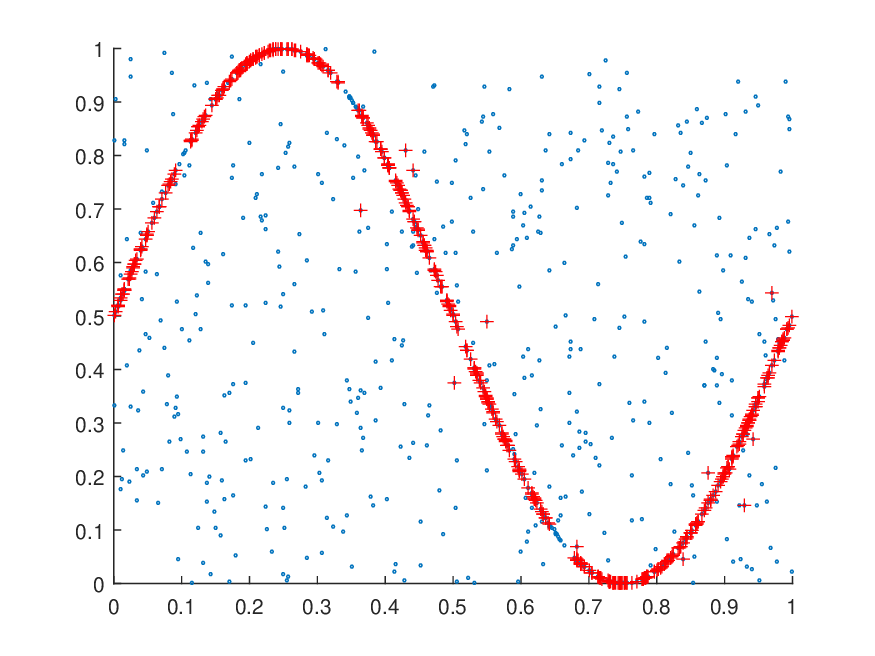}
			\label{datanonMres}
			
		\end{minipage}
		\vspace{0.05\linewidth}
		
	\end{minipage}
	\caption{Left column shows two-dimensional scatter plots for two pairs of features. The right column shows the scatter plot for the same data where the related observations identified by the proposed method are marked in red. Observations have a uniform distribution on $(0,1)$ on X axis. On Y axis the related observations are given by functions $y=4x^2$ and $y=0.5sin (2 \pi x)+0.5$  for top and bottom figures respectively. Remaining observations on Y axis have a uniform distribution on $(0,1)$}
	\label{figVarNM}
\end{figure}

The working of the proposed scheme has been shown, using two toy examples in Figure \ref{figVarNM}. The top figure in the left column shows the scatter plot for two variables. For a subset of data, the two features are related by the function $f(x)=4x^2$, where $x$ lies in the range $(0,1)$. For the rest of the observations, both variables are unrelated and have a uniform distribution in $(0,1)$. If the density of each cell is compared to the average density of the scatter plot, the unrelated observations will also get identified as biclusters as they are dense in comparison to the empty region lying in the upper part of the scatter plot. However, the proposed method can identify the observations related by the square function, as it compares the density of each cell with marginal density, as can be seen from the top image in the right column.

Similarly, the bottom left image gives a scatter plot for two variables. For a subset of data, the two variables are related by the function $f(x)=0.5\sin(2\pi x)+0.5$, where $x$ lies in the range $(0,1)$. For the rest of the observations, both variables are unrelated and have a uniform distribution in $(0,1)$. From the bottom image in the right column, we can see that the proposed method can identify the set of related observations (marked in red).  This is a non-monotonous relation, so UniBic will not be able to detect biclusters based on such relations. The relationships used in the first and the second images have been respectively used in generating simulated datasets for which results have been reported in the first and the second rows in Table \ref{accures}. These images have been presented here to clarify the motivation behind the proposed method.

\subsection{Choosing parameters for density estimates and convergence}
\label{choiceBinPrama}
The performance of histogram-based density methods is dependent on appropriate bin length (size of partitions used for density estimation). If the bin length reduces too quickly, the estimated density will be incorrect as some of the bins (partitions) may remain empty. On the other hand, if bin length decreases too slowly it will not converge to true density as it will be averaging over a large area. \cite{parzen} has given the criterion for convergence of density estimates as follows: The area of each bin $A_N$ should converge to zero as the number of observations $N$ goes to infinity. Also, the product of the number of observations $N$ and the area of each cell should go to infinity as $N$ goes to infinity. We can write the condition as $\lim_{N \to \infty} A_N \to 0$ and the product $ \lim_{N \to \infty} (N \times A_N )\to \infty$.

In this article, we use the separation distance based bin length. As discussed in Subsection \ref{smallHighDense}, for small datasets, we use the bin lengths $x_{len}=s_x^c$ and $y_{len}=s_y^c$ where $s_x$ and $s_y$ are maximal separations along $X$ and $Y$ axes, respectively and $c$ is the constant which should have a value close to $0.5$, but also less than $0.5$. We will now discuss how this choice leads to consistent estimates.

Suppose $N$ observations are distributed on the interval $[0,1]$ along a particular axis and the maximal separation along this axis be given by $s$. The work done by \cite{Deheuvels} shows that if certain regularity conditions are fulfilled, and probability density fulfils following conditions: 1) probability density function is defined on a bounded interval and 2) probability density has a non-zero minima on this interval, then limit inferior and superior are given by $\varliminf_{N \to \infty}s=C_1\log(N)/N$ and $\varlimsup_{N \to \infty}s=C_2\log(N)/N$, where $C_1$ and $C_2$ are different constants.

\begin{figure}[!htbp]
  	\centering

  	\begin{minipage}[t]{0.45\linewidth}

  			\includegraphics[width=\linewidth]{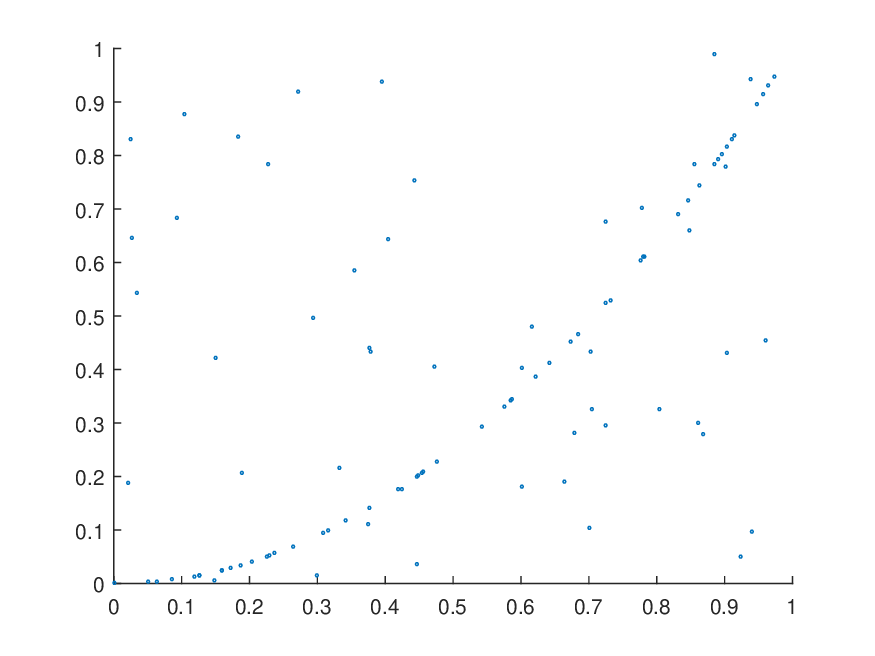}
  			\label{sparsedataset}

  	\end{minipage}
  	\hspace{0.05\linewidth}
  	\begin{minipage}[t]{0.45\linewidth}

  			\includegraphics[width=\linewidth]{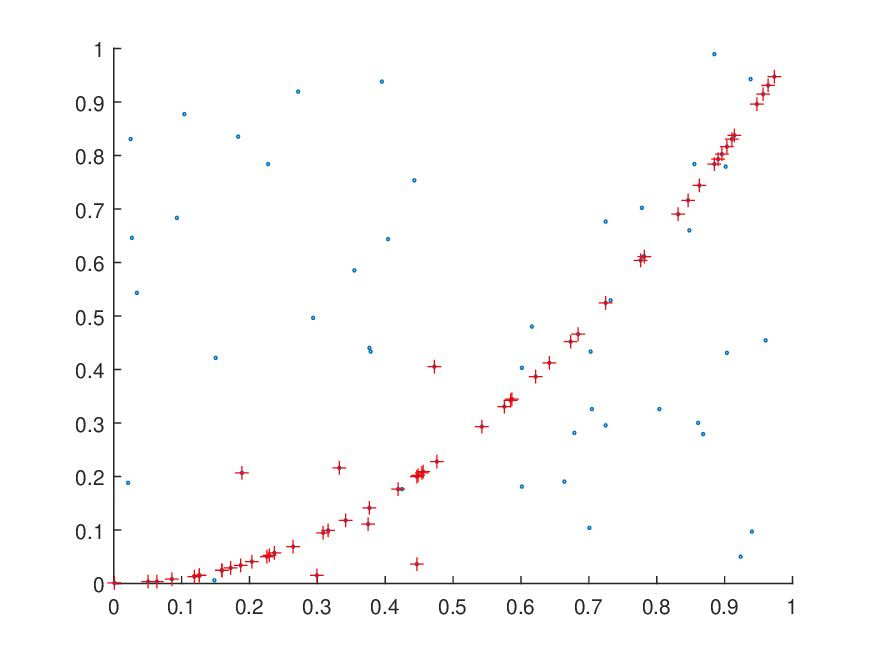}

  	\end{minipage}

  	\begin{minipage}[t]{0.45\linewidth}

  			\includegraphics[width=\linewidth]{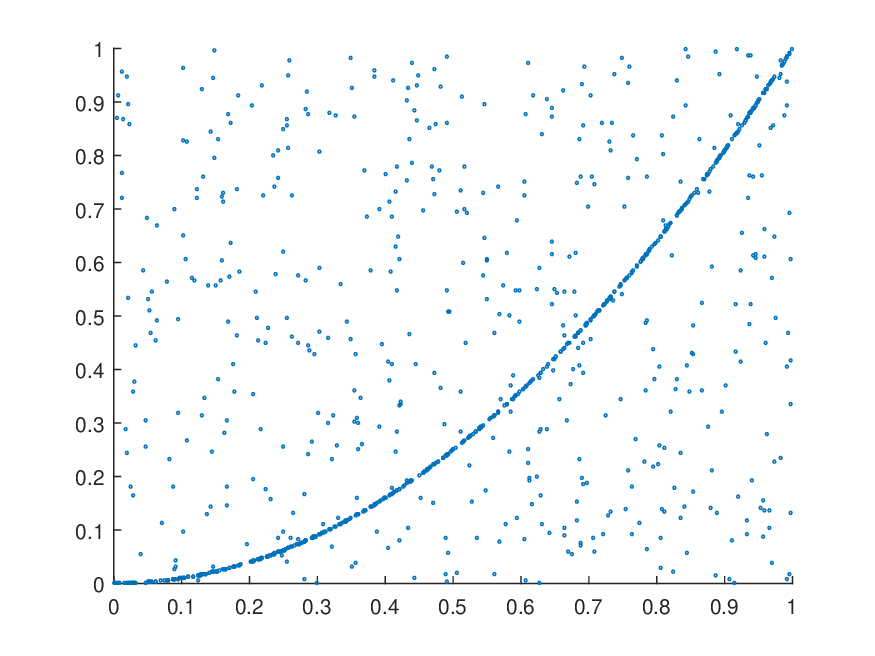}
  			\label{densedata}

  	\end{minipage}
  	\hspace{0.05\linewidth}
  	\begin{minipage}[t]{0.45\linewidth}

  			\includegraphics[width=\linewidth]{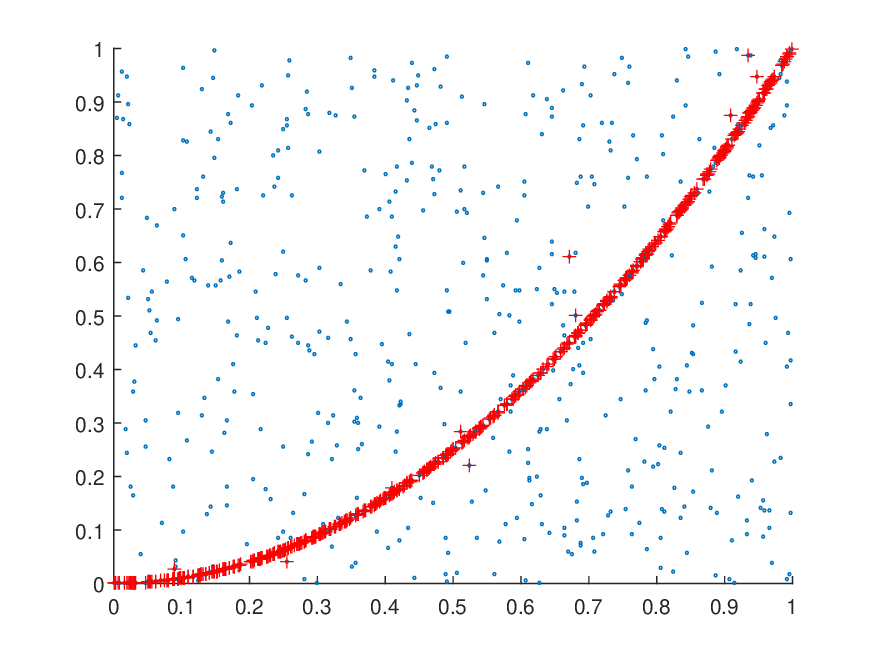}

  	\end{minipage}
  	\vspace{0.05\linewidth}
  	
  	\caption{Performance of the proposed method on small and large datasets. The left column shows all data in blue. The right column shows related observations identified by the proposed method in red and remaining observations in blue. Observations have a uniform distribution on $(0,1)$ on X axis. On Y axis the related observations are given by functions $y=x^2$.}
  	\label{figConsis}
  \end{figure}

For small datasets, the area of each cell is given by $A_N=s_x^c \times s_y ^c$. Since $c<0.5$, we see that upper limit is given by:
\begin{equation}
\varlimsup_{N \to \infty} A_N= \lim_{N \to \infty} (C_2\log(N)/N)^{2c}=0
\end{equation}
Parzen's second condition is also fulfilled as product of area and number of observations goes to infinity. We find that lower limit is given by:
\begin{equation}
\begin{split}
\varliminf_{N \to \infty} N \times A_N &= \lim_{N \to \infty} N \times (C_1\log(N)/N)^{2c}\\
&=\lim_{N \to \infty} N^{(1-2c)} (C_1\log(N))^{2c}=\infty\\
&\quad \text{as    } 1-2c>0
\end{split}
\end{equation}
Thus, estimated density for small datasets is consistent, according to Parzen density estimates.
For larger datasets we simply divide the $[0,1]$ interval along each axis into $3 \log (N) $ partitions. Thus, for a two-dimensional space normalized to $[0,1]\times[0,1]$ the area of each cell is $\lim_{N \to \infty}\frac{1}{9 \log^2(N)}$. Note that, $\lim_{N \to \infty}\frac{1}{9 \log^2(N)} = 0$ and $\lim_{N \to \infty}\frac{N}{9 \log^2(N)} = \infty$. Thus, we obtain consistent density estimates according to Parzen's rules. Figure \ref{figConsis} shows the performance of the proposed method on a small dataset containing only 100 observations and a larger dataset containing 1000 observations. The relation $f(x)=x^2$, where $x$ lies in $(0,1)$, exists for a subset of observations in both cases. We can see that the proposed method can find the related set of observations in both cases, as seen by observations marked in red.

\section{Stepwise Description of RelDenClu}
\label{detAlgo}
The previous section discussed the novel part of RelDenClu, while in this section we present the entire biclustering procedure. While CBSC by \cite{CBSC2018} compares the density of  a cell to average density, RelDenClu compares the cell density to marginal densities, as discussed in Section \ref{novAlgo}. 
Both algorithms use similar preprocessing and a similar procedure to obtain biclusters from dense related regions. Many algorithms like those proposed by \cite{FABIA2010}, \cite{subclu}, \cite{FIRESKriegel2005} use the apriori method to obtain biclusters from information obtained in low dimensional spaces. The procedure used by RelDenClu and CBSC is given in this section and it is different from the apriori method.

\subsection{Data preprocessing}
\label{prePro}
For finding dense regions, we need to normalize the data to $[0,1] \times [0,1]$. We use a common procedure to normalize the data, which has also been used by \cite{CBSC2018}. It is to be mentioned that, the data may come from a distribution with an unbounded base or a bounded base with an unknown range. As mentioned earlier, let data be represented by matrix $D$. The element in the $i_{th}$ row and the $j_{th}$ column is denoted by $e_{i,j}$. To normalize data from different distributions to the range $[0,1]$ we have applied one of the following two transformations to data:

\begin{equation}
norm_1(e_{i,j})=\frac{e_{i,j}-\min\limits_{1 \le k \le n}{(e_{k,j})}}{\max\limits_{1 \le k \le n}{(e_{k,j})}-\min\limits_{1 \le k \le n}{(e_{k,j})}}
\end{equation}

\begin{equation}
norm_2(e_{i,j})=norm_1( \tan ^{-1}(e_{i,j})/\pi +0.5)
\end{equation}

The transformation $norm_1(x)$ ($x$ represents elements of input data matrix) is commonly used for mapping data from a bounded base to $[0,1]$ interval while, transformation $norm_2(x)$ maps the interval $(-\infty, \infty)$ to interval $(0,1)$, which is a subset of $[0,1]$. The latter transformation is capable of mapping points from an unbounded region to a bounded interval. Thus it can be applied preferably if data follows distributions like Gaussian, which have non-compact base.
\subsection{Finding dense regions in two-dimensional space}
\label{denseregions}

In this section we will discuss in brief the procedure of finding dense regions. The data is normalized and dense regions are obtained using rolling window estimates or histogram-based estimates depending on the size of data. As mentioned earlier, a cell is called dense if its density is higher than the densities of the corresponding marginal cells. When the number of observations in the dataset is small, cell centered around each observation is considered. Overlapping cells are merged if the common region belonging to both the cells, has a density higher than the average density of marginal cells. For large datasets, a grid of disjoint cells is taken and neighbouring dense cells are merged to obtain larger dense regions. This step is already discussed elaborately in Section \ref{proposedHighDense} and the pseudo-code is given as Algorithm \ref{algodense}.
\begin{algorithm*}
	\caption{DenseRegions()}
	\label{algodense}
	\tcc{----------------------------------------------------}
	\tcc{This is an algorithm for finding dense regions in \\
		two-dimensional space defined by dimensions $i$ and $j$,\\
		for small dataset method described in Section \ref{smallHighDense}\\
		Input is list of all observations in two-dimensional
		space defined by $i$ and $j$		
	}
	\tcc{-----------------------------------------------------}
	\BlankLine
	
	\tcc{$sep_i$ is the maximum separation in dimension $i$}
	
	\For {each point $p_a$} {
		\tcc{$neigh(p_a)$ is the set of points representing neighbourhood of $p_a$}
		\tcc{Initialize neighbourhood of each point}
		$neigh(p_a)=\emptyset$
	}
	
	\For {each pair of points $p_a$, $p_b$} {
		\tcc{$d_i$ denotes distance in dimension $i$ and $d_j$ denotes distance in dimension $j$}
		\If {( $d_i(p_a, p_b)  < sep_i$ and $d_j(p_a, p_b)  < sep_j$)} {
			$connection(p_a, p_b)=1$\\
			Add $p_b$ to $neigh(p_a)$\\
			Add $p_a$ to $neigh(p_b)$\\ 
			
		}
		\Else {$connection(p_a, p_b)=0$}

	}

	$finset=\emptyset$\\

	\For {each point $p_a$} {
		\tcc{$MarginalNeigh_i(p_a)$ is set of points lying in the region given by $(p_a-sep_i/2, p_a+sep_i/2)$ in dimension $i$ and $(0,1)$ in dimension j}
		\tcc{$\#$ denotes cardinality of a set}
		\tcc{Here, $n$ is the `number of observations in dataset'}
		\If  {$\#(neigh(p_a))/(sep_i*sep_j)> n$ \textbf{and}
			$\#(neigh(p_a))/(sep_i*sep_j)> \#(MarginalNeigh_i(p_a))/sep_i$ \textbf{and} 					$\#(neigh(p_a))/(sep_i*sep_j)> \#(MarginalNeigh_j(p_a))/sep_j$}{
			$finset= finset \cup p_a$
		}
	}

	\For {each pair of points $p_a$, $p_b$ in $finset$} {
		
		\If {$\#(neigh(p_a) \cap neigh(p_b))*4/(sep_i *sep_j) >max(MarginalNeigh_i(p_a)/sep_i, MarginalNeigh_j(p_a)/sep_j, MarginalNeigh_i(p_b)/sep_i, MarginalNeigh_j(p_b)/sep_j )$ } {
			\tcc{Merge the two cells i.e., cell connected to one is automatically connected to other} 
			\tcc{$connection(p_a,:)$ denotes all connections of $p_a$}
			$connection(p_a, :)=connection(p_a,:)$ $\cup$ $connection(p_b,:)$\\
			$connection(p_b,:)=connection(p_a,:)$
		}

	}
	
	\tcc{Perform connected components according to new connections}
	\While{new connections are formed}{
		\For {each pair of points $p_a$, $p_b$ connected to each other}{
			$connection(p_a, :)=connection(p_a,:) \cup connection(p_b,:)$\\
			$connection(p_b,:)=connection(p_a,:)$
		}
	}

	\tcc{Retain only those points from $finset$ which have merged with other points}
	$finset$=$finset[sum(connection(p_a,:))>0]$
	
	$npc=\#(finset)$ 
	
	Find connected components in $finset$ using radius $rad= \log(npc)/npc$ \\
	
	\tcc{$k^{th}$ connected component is denoted by $Pts(i,j, k)$}
	\tcc{Let total number of connected components for given pair of dimensions be $m_{cc}$}
	
	Return points $<Pts(f_i, f_j, 1), Pts(f_i, f_j, 2) \cdots Pts(f_i, f_j, m_{cc})>$
\end{algorithm*}

\begin{algorithm*}
	\label{algobiclu}
	\caption{GetBiClusters}
	\tcc{This algorithm gives the process of obtaining biclusters from dense regions obtained by Algorithm \ref{algodense} }
	\tcc{The description of this algorithm is provided in Section \ref{mergebase}}
	\tcc{All the input parameters except data matrix $\mathcal{D}$ are described in Section \ref{effectparam}}
	\tcc{Let the data be $\mathcal{D}$}
	\tcc{Normalize the data using $norm_1$ if data seems to be \\
		coming from	a bounded distribution otherwise use $norm_2$}
	Normalize data using the procedure given in Section \ref{prePro}\\
	\tcc{Each pair of dimensions may contain several sets representing dense regions}
	\For {each pair of features $<f_i, f_j>$}
	{
		\tcc{This is described in Section \ref{smallHighDense} }
		${Pts(f_i, f_j, 1), Pts(f_i, f_j, 2), \cdots}$=denseregions($\mathcal{D}_{fi}$, $\mathcal{D}_{fj}$)
	} 
	
	\tcc{Initialize list of seed biclusters as given in the first paragraph of Section \ref{mergebase}}
	$Tlist=\emptyset$
	$Flist=\emptyset$
	\For {each set of 3 features $f_i, f_j, f_k $ and each combination of $u$, $v$, $w$} 
	{
		\tcc{Check if a seed bicluster is formed by each dense region for a set of three features }
		\tcc{dense regions numbered $u$, $v$ and $w$ in dimension pairs $<f_i, f_j>$,  $<f_j, f_k>$ \\
			and $<f_k, f_i>$ are checked for each combination between dense regions}
		
		$T_x=Pts(f_i, f_j, u)\cap Pts(f_j, f_k, v)\cap Pts(f_k, f_i,w )$\\
		\If {length($T_x<\mathtt{MinSeedSize}$)\label{algoweedparam}} 
		{Delete ($T_x$)} 
		\Else {
			$F_x=<u,v,w>$\\
			Add $T_x$ to Tlist\\
			Add $F_x$ to Flist
		}
	}
	Sort $Tlist$ in descending order of length\\

	\tcc{Remaining part of algorithm corresponds to Section \ref{mergebase} starting from second paragraph}
	$Clusterlist=\emptyset$
	
	\For {each $T_x$ in Tlist not marked as ignore} {
		$Clf=F_x$\\
		$SelectedTlist=T_x$\\

		\For {each $T_y$ in Tlist } {\label{addRows}
			\If {$(Clf \bigcap F_y) \neq \emptyset \wedge  length(T_x \bigcap T_y) > \mathtt{Sim2Seed} \times length(T_x)$}
			{
				$Clf=Clf \cup F_y$ 
			}
			
			\If {$\mathtt{ReuseAllSeeds}==FALSE \bigwedge length( T_x \bigcap T_y) > \mathtt{ReuseSeedSim} \times \mathtt{Sim2Seed} \times length (T_x)$}
			{ 
				\tcc{This seed bicluster will not be used as base}
				Mark $T_y$ as ignore  	
			}
		}
		Goto \ref{addRows} if seed biclusters have been added

		$Cl$ is list of observations which occur in at least $\mathtt{ObsInMinBase}$ bases present in $SelectedTlist$\\
		\If {$Cl \ne \emptyset$} {$Clusterlist=Clusterlist \text{ } \cup <Cl, Clf>$ }
		
	}
	
	\For {each pair of biclusters in Clusterlist $<Cl_i, Clf_i>,<Cl_j, Clf_j>$ } {
		\tcc{Formula for cosine similarity is given in Section \ref{mergebase}}
		\If{Cosine Similarity between $<Cl_i, Clf_i>,<Cl_j, Clf_j>>\mathtt{ClusSim}$ }
		{Delete the smaller bicluster}
	}
\end{algorithm*}
\subsection{Obtaining biclusters from dense regions found with different dimension pairs}
\label{mergebase}
Once we have obtained dense regions in two-dimensional spaces, this information is used to obtain biclusters in higher dimensions. The pseudo-code for the procedure described in this section is given as Algorithm \ref{algobiclu}.
\subsubsection{Finding seed biclusters}
\label{remnoise}
To identify the set of observations related to each other, we find relative density-based subsets in two-dimensional Euclidean space given by each pair of features. However, because the background noise can be distributed in arbitrary ways, this procedure of finding relations can result in a lot of noise. Therefore, to weed out noise, we search for a set of three features, for which a given set of observations form dense regions for each pair of features belonging to the set. Thus for a set of features $\{f_1, f_2, f_3\}$, we find dense regions for pairs $<f_1, f_2>$, $<f_2, f_3>$, and $<f_3, f_1>$. The intersection of observations forming dense regions for these three pairs is found. If the resulting set of observations is significant i.e., the number of observations is larger than a user-defined parameter named  $\mathtt{ObsInMinBase}$, we consider it to be a seed bicluster. The guideline to choose the values of  $\mathtt{ObsInMinBase}$ and other user-defined parameters is mentioned in Section \ref{effectparam}. For clarity, we have used the typewriter font for all such parameters. Once the set of observations has been identified, we treat it as a seed bicluster and add other feature-triplets which have significant overlap in their corresponding observation subsets. The following subsection describes the procedure for doing this.

\subsubsection{Using seed biclusters to get larger biclusters}
\label{getlargebi}
We first arrange seed biclusters in descending order by the number of observations present in each seed bicluster. We iteratively use these seed biclusters as a base for finding biclusters. When we use a particular seed bicluster for building a larger bicluster, we call it base seed bicluster. We start by initializing a bicluster $B$ to base seed bicluster $S$. We compare this bicluster $B$ with all other seed biclusters (except $S$). If the length of $S$ in terms of the number of observations is denoted by $length(S)$, we find seed biclusters which have at least $length(S)\times \mathtt{Sim2Seed}$ observations common with $B$, where $\mathtt{Sim2Seed}$ is a user-defined parameter. As we keep finding seed bicluster matching with $B$, we update $B$ by adding all observations present in matching seed bicluster. This is done repeatedly till no more additions can be made to $B$. Then, we find observations which occur in at least $\mathtt{ObsInMinBase}$ number of seed biclusters, where $\mathtt{ObsInMinBase}$ is a user-defined parameter. These observations and features corresponding to the matching seed biclusters form an output bicluster.

The user also has an option of ignoring seed biclusters which have very high overlap with some other seed bicluster already used as the base. To use this option one must set the user-defined parameter $\mathtt{ReuseAllSeeds}$ to false. In this case, the user-defined parameter named $\mathtt{ReuseSeedSim}$ gives the threshold for the amount of overlap between seed biclusters. If the overlap between a base seed bicluster $S$ and other seed bicluster is greater than $\mathtt{ReuseSeedSim}\times \mathtt{Sim2Seed} \times length(S)$, we do not use the matching seed bicluster as base seed bicluster in future iterations.
On the other hand, if the parameter  $\mathtt{ReuseAllSeeds}$ is set to true we will not ignore seed biclusters and they will be used as base in subsequent iterations. These two parameters control the amount of overlap between biclusters in terms of observations.

\subsubsection{Weeding out similar biclusters}
The parameter $\mathtt{ClusSim}$ is used to weed out overlapping biclusters, as it controls the maximum amount of similarity allowed between biclusters. For  each pair of biclusters we calculate the similarity between them using the formula: $\frac{\#(O_1 \bigcap O_2)}{\sqrt{\#(O_1)}\times\sqrt{\#(O_2)}}\times \frac{\#(F_1 \bigcap F_2)}{\sqrt{\#(F_1)}\times\sqrt{\#(F_2)}}$. Here $O_1$ and $O_2$ denote the sets of observations in the first and the second biclusters, respectively. Similarly, $F_1$ and $F_2$, respectively denote the sets of features in the first and the second biclusters. $\#()$ is used to denote the cardinality of a set. If the similarity between the two biclusters is found to be greater than $\mathtt{ClusSim}$, we discard the smaller ones.

\subsection{Choice of parameters for the proposed method}
\label{effectparam}
In this section, we provide a guideline to choose the range of each parameter. The values of various parameters used for different experiments are reported in Section \ref{resSec}.

$\mathtt{MinSeedSize}$: While choosing seed biclusters in Step \ref{algoweedparam} in Algorithm \ref{algobiclu}, this parameter is used. We make a list of all seed biclusters longer than $\mathtt{MinSeedSize}$ and use only these biclusters for further processing. Other seed biclusters are deleted. Setting lower values will allow the algorithm to detect biclusters with fewer observations. Setting a larger value will leave out smaller biclusters and report only large ones. If we want all small and large biclusters, it can be set to three, as we have defined connectedness based common dense region for three pairs in Section \ref{mergebase}.

$\mathtt{Sim2Seed}$: In Section \ref{mergebase}, a base seed bicluster is compared to other seed biclusters to see if the latter can be included in bicluster, based on the similarity between the two. This similarity between base seed bicluster $S$ and other seed biclusters $S'$ is judged using the number of observations present in both $S$ and $S'$. If this number is greater than $\mathtt{Sim2Seed} \times length(S)$, they are said to be similar. The range $(0.6, 0.98)$ is seen to give stable results.

$\mathtt{ReuseAllSeeds}$: Each seed bicluster can be used as a base, for finding a bicluster. For using each seed bicluster as a base this parameter should be set to $\mathtt{TRUE}$. Otherwise, seeds previously found to have a high overlap with base seed bicluster are not used as bases in further calculations.

$\mathtt{ReuseSeedSim}$: If we ignore some seed biclusters while choosing the bases, we set 
$\mathtt{ReuseAllSeeds}$ to $\mathtt{FALSE}$. In this case, we need to set the parameter $\mathtt{ReuseSeedSim}$. If the number of common observations between base seed bicluster $S$ and other seed bicluster $S'$ is greater than $\mathtt{ReuseSeedSim} \times \mathtt{Sim2Seed} \times length(S)$, $S'$ is not used as base. The value of this parameter lies in $(0,1)$. Setting a higher value of this parameter allows the algorithm to detect biclusters with a greater amount of overlap. If the parameter is set to a lower value, less overlap occurs and computation time is also less.

$\mathtt{ObsInMinBase}$: If an observation is present in at least $\mathtt{ObsInMinBase}$ number of dense regions forming a bicluster, it is included in the set of observations corresponding to the bicluster. In our experiments, we have used values in range 3 to 15.

$\mathtt{ClusSim}$: If cosine similarity defined in Section \ref{mergebase}, between biclusters is found to be larger than $\mathtt{ClusSim}$, smaller bicluster is ignored. Unlike $\mathtt{ReuseSeedSim}$, here we compare biclusters in terms of both the number of observations and the number of features. Also, it weeds out biclusters once they are found and thus provides finer control.


\section{Methodology for Experimental results}
\label{methSec}
To show the effectiveness of the proposed algorithm we have generated several simulated datasets, as described in Section \ref{secgendata} and appendix \ref{artDatDet}.The usefulness of the proposed algorithm is demonstrated using real-life datasets. The results for simulated and real-life datasets are reported in Section \ref{resSec}. Section \ref{relexec} reports the execution time needed for the proposed algorithm and other algorithms for various simulated datasets. In our experiments, we have considered a dataset to be small if it has less than 750 observations, else it is considered to be large.  

\subsection{Algorithms used for comparison}
\label{algoused}
The performance of the proposed algorithm is compared with seven other algorithms (which are described below in brief) namely CLIQUE, Proclus, ITL, Subclu, P3C, UniBic, CBSC for both simulated and real-life datasets. For CBSC and UniBic,  implementations, as provided by the respective authors, have been used. For CBSC, we have used same parameter values as used by \cite{CBSC2018}. For UniBic, the default values of parameters have been used. For all other algorithms, implementations available in R package named subspace by \cite{subspaceRpack} have been used, with the parameters given in the help page.
\begin{itemize}
	\item CLIQUE: This algorithm by \cite{CLIQUE1998} uses a grid-based approach to find dense regions in low dimension space. Candidate biclusters generated are analyzed in higher dimensions after adding features gradually. Thus, it uses the monotonicity property of high-density regions i.e., a set of points forming a high-density region in high dimensional space must form a high-density region in lower dimensions.
	
	\item Proclus: This algorithm by \cite{PROCLUS}, uses a k-medoid like approach for identifying biclusters in different subspaces. The biclusters obtained tend to be spherical. It is known to be a robust biclustering technique.
	
	\item ITL: The objective of this algorithm, proposed by \cite{ITL2003}, is to minimize within-cluster divergence and maximize between-cluster divergence for features.
	
	\item Subclu: Subclu by \cite{subclu} is an example of a procedure where density connected regions form a bicluster. A region is said to be dense if it is composed of connected open discs of radius $r$ each having at least $\epsilon$ observations. Thus density becomes the effective criterion function and its threshold value is decided by $\epsilon$ and $r$. Dense regions found in smaller subspaces are combined using the apriori approach, and density techniques are used again to obtain actual biclusters from the set of candidate biclusters.
	
	\item P3C: This is also a grid-based method proposed by \cite{P3C2008}, which identifies dense cells using a statistical approach. Once probable biclusters are generated in lower dimension space, the expectation-maximization algorithm is used. Initially, the observations are assigned using a fuzzy membership criterion.
	
	\item UniBic: This algorithm by \cite{unibic} finds trend preserving biclusters. The seed biclusters are identified using the Longest Common Subsequence framework. Rows are added to these seed biclusters to obtain final biclusters.
	
	\item CBSC: CBSC by \cite{CBSC2018} is a method where density estimation is done in two-dimensional space using windows of radius calculated using the length of the Minimal Spanning tree. This procedure implicitly identifies a sharp change in density in two-dimensional space. Since the radius is calculated separately for each pair of dimensions, the procedure takes care of specific density variations in each two-dimensional subspace. Afterwards, results from different two-dimensional spaces are combined.  
\end{itemize}
CLIQUE, P3C and Subclu have been chosen for comparison as they are density-based methods. Proclus has been chosen as it is robust in presence of noise. UniBic is a recent algorithm, which provides good results in comparison to many existing methods like FABIA by \cite{FABIA2010} and ISA by \cite{ISA}. ITL finds biclusters based on feature similarity using mutual information. CBSC also searches for features leading to relation-based biclusters using a density-based approach.

\subsection{Generating simulated datasets}
\label{secgendata}

To check whether the proposed biclustering algorithm can detect biclusters based on linear and non-linear relationships, we have experimented with several simulated datasets. We also want to see the effect of applying different data transforms on the performance of the proposed algorithm. Each type of dataset corresponds to one row in Table \ref{accures}, which reports the mean and standard deviation of accuracy obtained by different algorithms (where for each type of dataset, ten dataset instances are used). The first two rows (``Non-linear 1'' and ``Non-linear 2'') of Table \ref{accures} represent datasets having biclusters based on non-linear relations. The third row (``Base'') represents dataset containing bicluster based on linear relation. Following nine rows (``Scaled'', ``Translated'', ..., ``Permutations'') give results for datasets obtained by applying different transforms to the ``Base'' dataset. These datasets allow us to analyze the behaviour of proposed method in a way similar to analysis done by \cite{addmissClu1971} for clustering algorithms. Next two rows (``Normal'' and ``Noisy Normal'') contain results for datasets generated using normal distribution with and without noise. The last two rows (``Overlap 1'' and ``Overlap 2'') give results for two overlapping biclusters existing within a dataset. The method for generating the datasets for each row is discussed in the appendix \ref{artDatDet}.

\subsection{Evaluating the performance on simulated datasets}
In this section, we discuss how the performance of proposed and seven other state-of-art algorithms are evaluated for above-mentioned simulated datasets. 
For data given in an $N \times M$ matrix, accuracy is calculated as follows: generate a matrix $m_B$ such that $m_B(i,j)=1$ if observation $i$ and feature $j$ is a member of actual bicluster in data. Similarly, generate a matrix $m_E$ for estimated bicluster. 

Accuracy is given by $sum(XNOR(m_B, m_E))/(N \times M)$ i.e. the ratio where $m_B$ matches $m_E$ to the size of the data matrix. For each bicluster present in data, the best match is reported for each algorithm.  Significance of the results is analyzed using pairwise right-tailed $\textit{t-test}$, and corresponding $\textit{p-values}$ are also reported.

\subsection{Real-life datasets used for comparison}
\label{realDs}
In this section we briefly describe the real-life datasets used for comparison (in Section \ref{realdatares}). 
\subsubsection{Real-life datasets used for discovering biclusters}
\label{cluRealds}
We have used three datasets from UCI ML repository by \cite{UCIML2019} to discover classes existing in the dataset as biclusters (reported in Section \ref{UnsupRes}). Description of these datasets are given below.
\begin{itemize}
	\item Magic: It is `MAGIC Gamma Telescope Data Set' of size $19020 \times 10$ and  contains observations representing gamma and hadron particles. 
	\item Cancer: It is known as `Breast Cancer Wisconsin (Original) Data Set' of size $683 \times 9$. The observations represent the conditions 'benign' and 'malignant'.
	\item Ionosphere: The size of `Ionosphere Data Set' is $351 \times 34$. The observations represent good and bad radar signals.
\end{itemize}
\subsubsection{Real-life datasets used for classification using additional features generated by biclustering methods}
\label{supRealds}
We have also used biclustering algorithms to improve the classification accuracy of three datasets described below (from UCI ML repository). The corresponding results are reported in Section \ref{supRes}.
\begin{itemize}
	\item Credit Card: The `Default of credit card clients Data Set' is of size $30000 \times 23$. It contains observations representing credit card defaulters and customers who make regular payments. 
	\item Image Segmentation: The `Image Segmentation Data Set' is of size $2310 \times 19$. It contains data from seven different classes. Each instance represents $3 \times 3$ pixel region, drawn randomly from a database of 7 outdoor images.
	\item Bioconcentration: It is named `QSAR Bioconcentration classes Data Set' and has a size of $779 \times 14$. It is a dataset of chemicals. The chemicals belong to three classes depending on whether a chemical: (1) is mainly stored within lipid tissues, (2) has additional storage sites (e.g. proteins), or (3) is eliminated. 
\end{itemize}

\section{Results}
\label{resSec}

\begin{table*}[!htbp]
	\caption{Accuracy on Simulated datasets (Mean and Standard Deviation for 10 datasets of each type)}
	\label{accures}
	\resizebox{\textwidth}{!}{
		\begin{tabular}{|c|c|c|c|c|c|c|c|c|c|}
			\hline
			\multirow{2}{*}{Dataset}&\multirow{2}{*}{Accuracy}	&\multicolumn{8}{c|}{Methods used}\\
			\cline{3-10}
			
			&&CLIQUE&Proclus&ITL&Subclu&P3C&UniBic&CBSC&Proposed 	\\			
			
			\hline
			\multirow{2}{*}{Non-Linear 1} &Mean&0.784&0.781&0.587&0.785&0.765&0.610&0.669&\textbf{0.913}\\
			&Deviation&0.002&0.014&0.024&0.017&0.004&0.142&0.049&0.022\\
			\hline
			\multirow{2}{*}{Non-Linear 2}&Mean&0.782&0.792&0.445&0.793&0.764&0.694&0.752&\textbf{0.883}\\
			&Deviation&0.002&0.007&0.119&0.016&0.003&0.107&0.022&0.006\\
			
			\hline
			\multirow{2}{*}{Base} &Mean&0.702&0.785&0.701&0.794&0.781&0.839&0.834&\textbf{0.989}\\
			&Deviation&0.247&0.013&0.355&0.023&0.04&0.034&0.063&0.006\\
			\hline
			\multirow{2}{*}{Scaled}&Mean&0.702&0.768&0.701&0.759&0.782&0.684&0.834&\textbf{0.989}\\
			&Deviation&0.247&0.015&0.355&0.016&0.04&0.211&0.063&0.006\\
			\hline
			\multirow{2}{*}{Translated}&Mean&0.702&0.803&0.701&0.788&0.779&0.582&0.834&\textbf{0.989}\\
			&Deviation&0.247&0.023&0.355&0.019&0.041&0.159&0.063&0.006\\
			\hline
			\multirow{2}{*}{Linear transform}&Mean&0.702&0.77&0.701&0.765&0.779&0.839&0.834&\textbf{0.989}\\
			&Deviation&0.247&0.028&0.355&0.025&0.041&0.034&0.063&0.006\\
			\hline
			
			\multirow{2}{*}{Square}&Mean&0.827&0.805&0.75&0.8&0.771&0.847&0.696&\textbf{0.981}\\
			&Deviation&0.018&0.021&0.021&0.024&0.009&0.038&0.15&0.005\\
			\hline
			\multirow{2}{*}{Exponential}&Mean&0.793&0.791&0.729&0.811&0.848&0.861&0.866&\textbf{0.978}\\
			&Deviation&0.026&0.016&0.023&0.025&0.055&0.046&0.062&0.020\\
			\hline
			\multirow{2}{*}{Point proportion}&Mean&0.702&0.8&0.701&0.802&0.787&0.804&0.834&\textbf{0.992}\\
			&Deviation&0.247&0.024&0.355&0.03&0.022&0.046&0.063&0.006\\
			\hline
			\multirow{2}{*}{Cluster proportion}&Mean&0.643&0.711&0.586&0.724&0.802&0.8&0.937&\textbf{0.996}\\
			&Deviation&0.227&0.024&0.029&0.02&0.092&0.078&0.02&0.003\\
			\hline
			\multirow{2}{*}{Noisy uniform}&Mean&0.696&0.792&0.71&0.795&0.772&0.592&0.851&\textbf{0.939}\\
			&Deviation&0.245&0.01&0.032&0.017&0.032&0.124&0.049&0.029\\
			\hline
			\multirow{2}{*}{Permutations}&Mean&0.695&0.785&0.6&0.798&0.783&0.839&0.834&\textbf{0.989}\\
			&Deviation&0.245&0.015&0.189&0.022&0.015&0.034&0.063&0.006\\
			\hline
			\multirow{2}{*}{Normal}&Mean&0.799&0.783&0.504&0.783&0.767&0.898&0.805&\textbf{0.991}\\
			&Deviation&0.011&0.017&0.013&0.01&0.017&0.025&0.075&0.003\\
			\hline
			\multirow{2}{*}{Noisy normal}&Mean&0.787&0.772&0.512&0.78&0.758&0.522&0.769&\textbf{0.901}\\
			&Deviation&0.015&0.011&0.011&0.015&0.047&0.098&0.074&0.031\\
			\hline
			\multirow{2}{*}{Overlap 1}&Mean&0.779&0.744&0.847&0.754&0.831&0.446&0.85&\textbf{0.963}\\
			&Deviation&0.031&0.005&0.005&0.014&0.007&0.064&0.087&0.062\\
			\hline
			\multirow{2}{*}{Overlap 2}&Mean&0.856&0.836&0.759&0.833&0.856&0.534&0.818&\textbf{0.975}\\
			&Deviation&0.025&0.008&0.086&0.006&0.021&0.054&0.098&0.033\\
			\hline

		\end{tabular}
	}
\end{table*}


To show the effectiveness of the proposed method, experiments have been conducted with simulated and real-life datasets mentioned earlier. Simulated datasets have been used to verify the performance of the proposed method on datasets with varying properties. Accuracy has been used as a performance metric for all datasets. Further, for simulated datasets paired $\textit{t-test}$ has been done and corresponding $\textit{p-values}$ have been reported to show the significance of the results (Tables \ref{accures}, \ref{pairedtest} and Figures \ref{barNL}, \ref{barTrans}, \ref{barNormal}, \ref{barOverlap} in section \ref{artRes}).

Three datasets namely Magic, Breast cancer and Ionosphere (described in Section \ref{cluRealds}), each containing two classes, have been used to compare the performance of the proposed and other biclustering algorithms. RelDenClu finds biclusters corresponding to known classes with greater accuracy (Tables \ref{magicTab}, \ref{cancerTab}, \ref{ionTab} and Figure \ref{barClus} in Section \ref{UnsupRes}). Note that, the label information has not been used. Along with accuracy, precision, recall \citep{FawcettROCPRL} and g-score (G-score is the geometric mean of precision and recall) have also been reported. 

As an application to supervised learning, the biclustering algorithms have been used to generate new features for three datasets namely Credit card, Image segmentation and Bioconcentration (as described in Section \ref{supRealds}).  From the results (Table \ref{tabresSup} and Figure \ref{barSup}) discussed in Section \ref{supRes}, it is found that RelDenClu provides greater improvement in classification accuracy compared to seven other methods for all the three datasets thereby leading to better identification of credit card defaulters, improved understanding of chemical bioconcentration in tissues, and better image segmentation results.

Execution time for simulated datasets have been provided in Section \ref{relexec} (Table \ref{runtimetab}), along with time complexity of the proposed method. Additionally, accuracies and execution times for a large simulated dataset have also been presented in Table \ref{exectimelarge}. In all the tables reporting the results, best accuracies are shown in bold.

\subsection{Results on simulated datasets}
\label{artRes}

\begin{table*}[!htbp]
	
	\caption{Paired $\textit{t-test}$ statistics  and $\textit{p-values}$ using simulated datasets, $\textit{p-values}<0.01$ shows that proposed method provides significantly better accuracy}
	\label{pairedtest}
	\resizebox{\textwidth}{!}{
		\begin{tabular}{|c|c|c|c|c|c|c|c|c|}
			
			\hline
			\multirow{2}{*}{Dataset}&\multirow{2}{*}{Statistics}	&\multicolumn{7}{c|}{Methods used for comparing with the proposed method}\\
			\cline{3-9}

			&&CLIQUE&Proclus&ITL&Subclu&P3C&UniBic&CBSC\\
			
			\hline
			\multirow{2}{*}{Non-Linear 1} &t& 18.71& 18.71& 30.90& 13.85& 19.81& 6.53& 13.26\\
			&$\textit{p-value}$& 8.15e-09& 8.17e-09& 9.53e-11& 1.13e-07& 4.94e-09& 5.36e-05& 1.64e-07\\
			\hline
			\multirow{2}{*}{Non-Linear 2}&t& 40.78& 24.34& 11.47& 15.64& 59.55& 5.55& 17.11\\
			&$\textit{p-value}$& 7.99e-12& 7.99e-10& 5.67e-07& 3.93e-08& 2.68e-13& 1.78e-04& 1.79e-08\\
			\hline
			
			\multirow{2}{*}{Base} &t& 3.71& 47.12& 25.65& 24.31& 16.74& 13.91& 7.76\\
			&$\textit{p-value}$& 2.43e-03& 2.19e-12& 5.02e-10& 8.07e-10& 2.17e-08& 1.09e-07& 1.42e-05\\
			\hline
			\multirow{2}{*}{Scaled}&t&3.71& 51.05& 25.65& 41.87& 16.72& 4.55& 7.76\\
			&$\textit{p-value}$&2.43e-03& 1.07e-12& 5.02e-10& 6.31e-12& 2.19e-08& 6.94e-04& 1.42e-05\\
			\hline
			\multirow{2}{*}{Translated}&t&3.71& 24.13& 25.65& 30.32& 16.66& 8.09& 7.76\\
			&$\textit{p-value}$&2.43e-03& 8.63e-10& 5.02e-10& 1.13e-10& 2.26e-08& 1.02e-05& 1.42e-05\\
			\hline
			\multirow{2}{*}{Linear transform}&t& 3.71& 23.78& 25.65& 26.74& 16.46& 13.91& 7.76\\
			&$\textit{p-value}$&2.43e-03& 9.83e-10& 5.02e-10& 3.46e-10& 2.51e-08& 1.09e-07& 1.42e-05\\
			\hline
			\multirow{2}{*}{Square}&t&27.79& 31.54& 34.39& 26.84& 56.71& 10.64& 6.03\\
			&$\textit{p-value}$&2.46e-10& 7.95e-11& 3.67e-11& 3.35e-10& 4.15e-13& 1.07e-06& 9.78e-05\\
			\hline
			\multirow{2}{*}{Exponential}&t&16.68& 21.10& 22.77& 15.16& 7.62& 6.23& 5.62\\
			&$\textit{p-value}$&2.24e-08& 2.83e-09& 1.44e-09& 5.14e-08& 1.62e-05& 7.69e-05& 1.63e-04\\
			\hline
			\multirow{2}{*}{Point proportion}&t&3.74& 23.64& 26.23& 19.28& 33.80& 13.80& 8.04\\
			&$\textit{p-value}$&2.32e-03& 1.03e-09& 4.12e-10& 6.27e-09& 4.29e-11& 1.16e-07& 1.06e-05\\
			\hline
			\multirow{2}{*}{Cluster proportion}&t&4.91& 37.66& 46.93& 45.52& 6.67& 8.13& 8.91\\
			&$\textit{p-value}$&4.15e-04& 1.63e-11& 2.27e-12& 2.98e-12& 4.60e-05& 9.76e-06& 4.64e-06\\
			\hline
			\multirow{2}{*}{Noisy uniform}&t&3.14& 13.47& 14.19& 11.52& 11.37& 8.06& 4.59\\
			&$\textit{p-value}$&5.97e-03& 1.43e-07& 9.11e-08& 5.44e-07& 6.09e-07& 1.04e-05& 6.55e-04\\
			\hline
			\multirow{2}{*}{Permutations}&t&3.84& 35.37& 6.53& 25.11& 34.30& 13.91& 7.76\\
			&$\textit{p-value}$&1.99e-03& 2.85e-11& 5.37e-05& 6.06e-10& 3.76e-11& 1.09e-07& 1.42e-05\\
			\hline
			\multirow{2}{*}{Normal}&t&50.63& 44.11& 110.85& 64.68& 38.73& 13.07& 7.82\\
			&$\textit{p-value}$&1.15e-12& 3.95e-12& 1.00e-15& 1.27e-13& 1.27e-11& 1.85e-07& 1.32e-05\\
			\hline
			\multirow{2}{*}{Noisy normal}&t&10.37& 14.11& 39.07& 9.72& 8.87& 10.11& 5.54\\
			&$\textit{p-value}$&1.32e-06& 9.61e-08& 1.17e-11& 2.26e-06& 4.79e-06& 1.63e-06& 1.80e-04\\
			\hline
			\multirow{2}{*}{Overlap 1}&t&9.30& 10.65& 5.85& 11.28& 6.45& 17.52& 2.83\\
			&$\textit{p-value}$1&3.27e-06& 1.06e-06& 1.22e-04& 6.53e-07& 5.91e-05& 1.46e-08& 9.79e-03\\
			\hline
			\multirow{2}{*}{Overlap 2}&t&11.66& 11.70& 8.07& 13.96& 9.24& 24.21& 4.43\\
			&$\textit{p-value}$&4.91e-07& 4.76e-07& 1.04e-05& 1.05e-07& 3.44e-06& 8.39e-10& 8.22e-04\\
			
			\hline

		\end{tabular}
	}
\end{table*}

\begin{figure}[!h]
	\caption{Accuracy obtained using various algorithms for Non-Linear datasets}
	\label{barNL}
	\includegraphics[width=\linewidth]{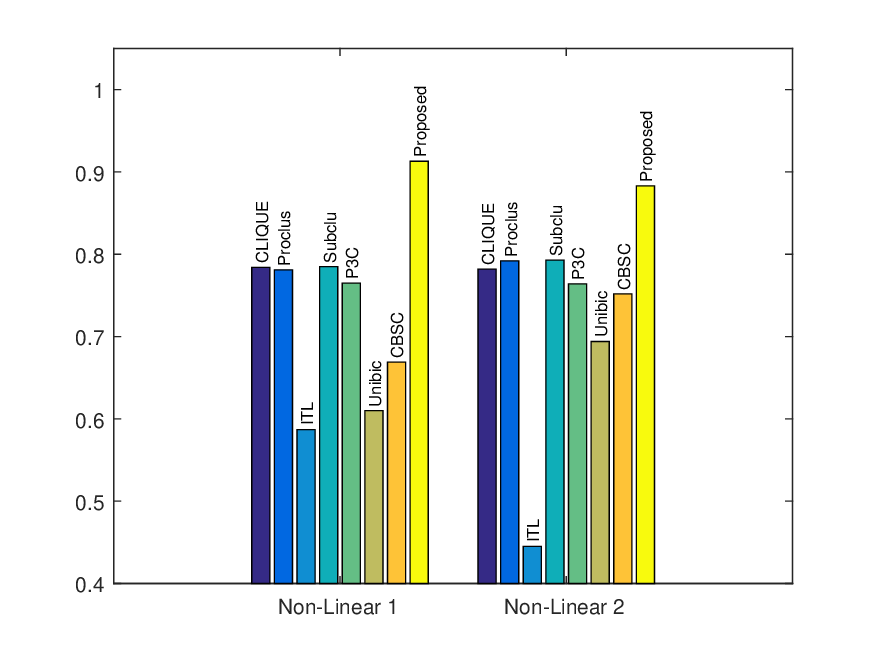}

\end{figure}

\begin{centering}
\begin{figure}
	\caption{Accuracy obtained using various algorithms for Base and transformed datasets}
	\label{barTrans}
	\includegraphics[width=1.7\linewidth,height=\linewidth, angle=270]{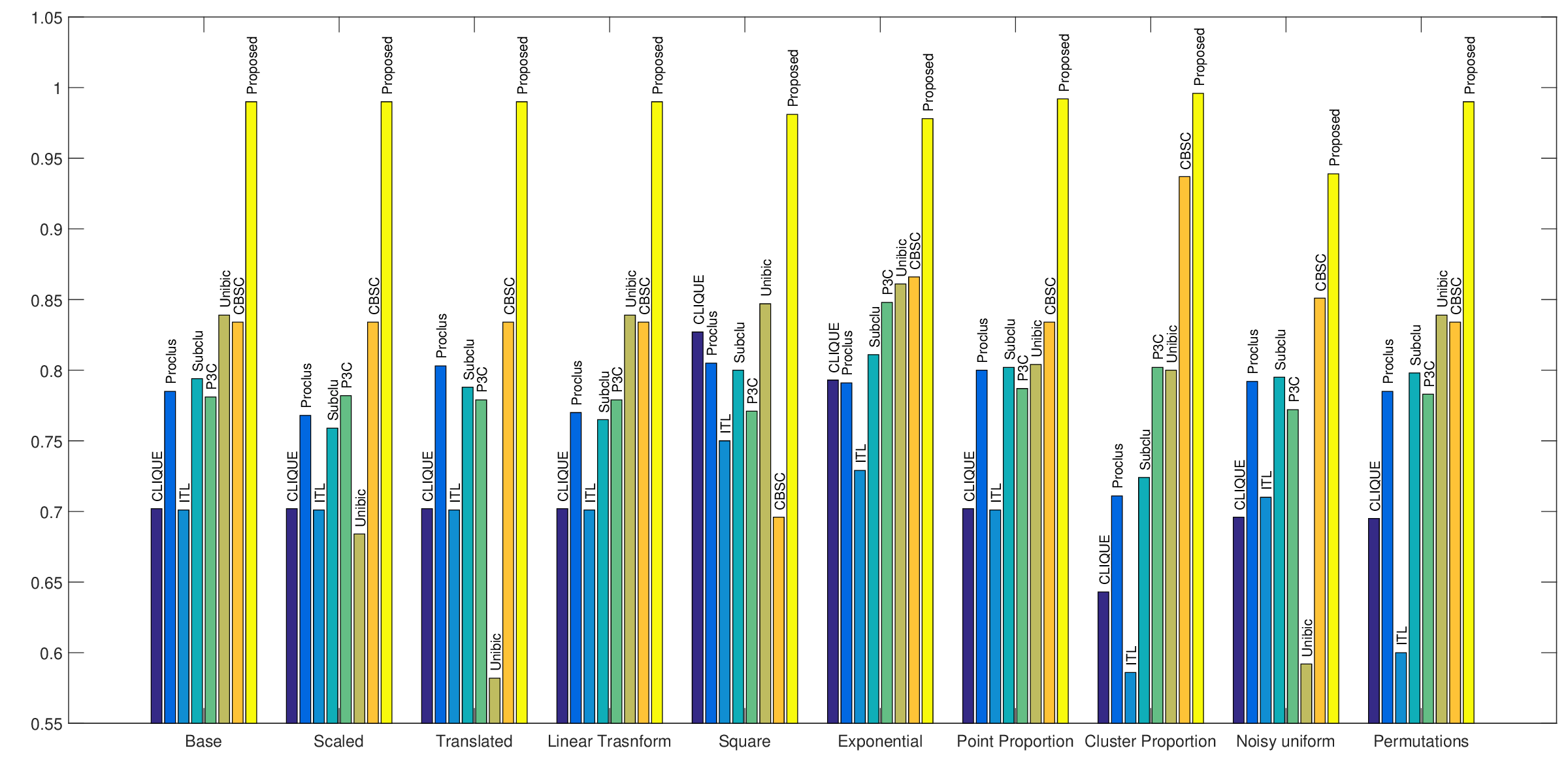}
\end{figure}
\end{centering}

 \begin{figure}[]
 	\caption{Accuracy obtained using various algorithms for Normal and Noisy Normal datasets}
 	\label{barNormal}
 	\includegraphics[width=\linewidth]{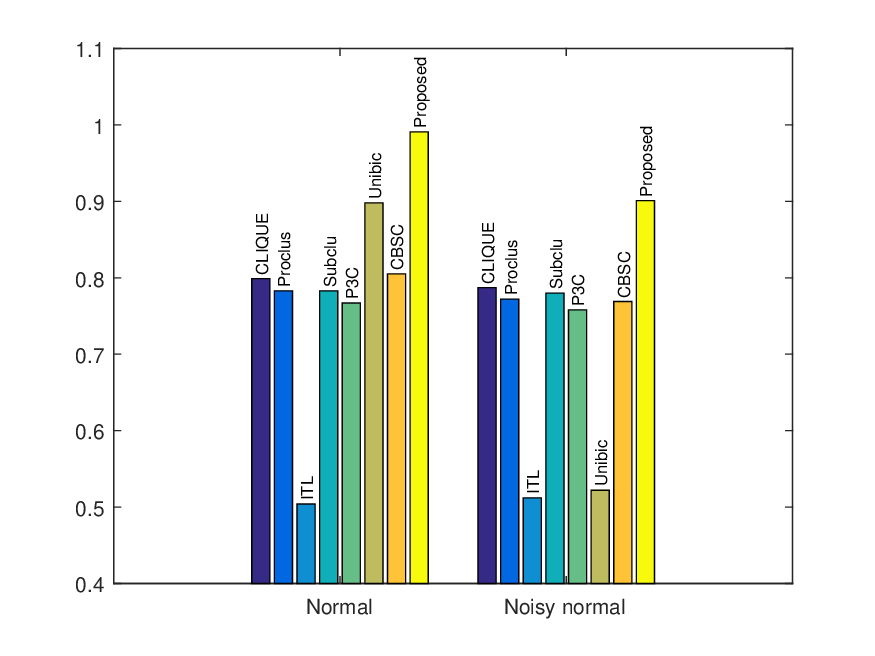}
 \end{figure}
 \begin{figure}[]
 	\caption{Accuracy obtained using various algorithms for dataset containing overlapping biclusters}
 	\label{barOverlap}
 	\includegraphics[width=\linewidth]{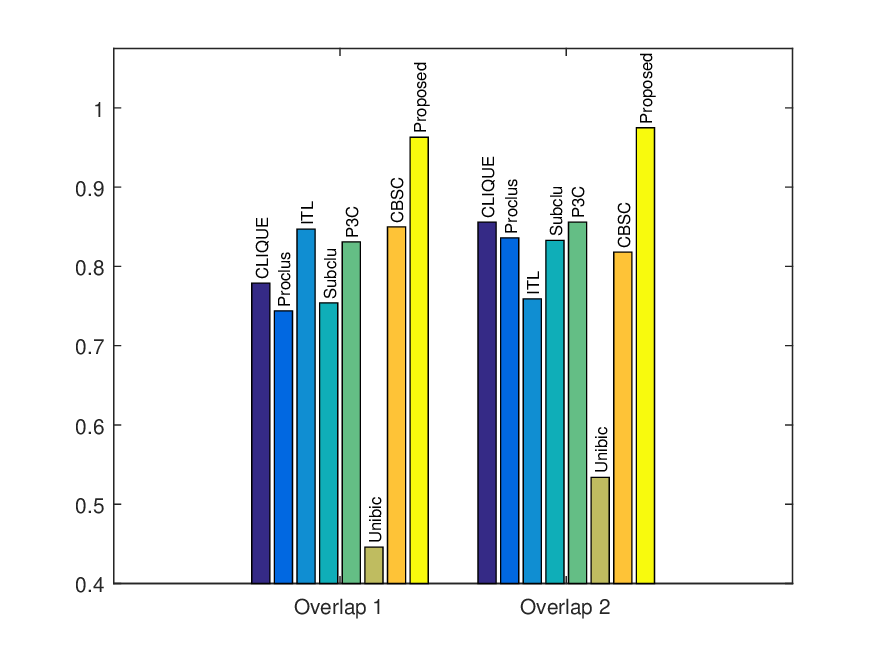}
 \end{figure}

For simulated data, experiments are conducted with 10 different instances and the average (over 10 simulations) accuracy and corresponding standard deviation (denoted as "deviation") values are put in Table \ref{accures}. We see that the proposed method provides better accuracy for all simulated datasets. To understand the significance of results in Table \ref{accures}, we have also reported the corresponding results of paired $\textit{t-test}$ along with $\textit{p-values}$ in Table \ref{pairedtest}. Here we also notice that the proposed method provides significantly better accuracy ($\textit{p-value}<0.01$) as compared to other methods. 

It is seen from Table \ref{accures}, that the proposed algorithm provides higher accuracy for two types of non-linear datasets (``Non-linear 1'' and ``Non-linear 2''). Unlike CBSC and other algorithms, the proposed method finds the biclusters accurately in-spite of highly variable distribution of background observations as seen from the row ``Non-Linear 1''. Its accuracy for datasets containing bicluster based on highly non-monotonous data is also higher in comparison with other algorithms as seen from row ``Non-Linear 2''. The values of accuracy obtained using different algorithms on these datasets are presented in Figure \ref{barNL} as bar graphs, for better illustration. 

From Figure \ref{barTrans} and Table \ref{accures}, we find that the proposed method has better accuracy for ``Base'' dataset and all its transforms , i.e. ``Scaled'',...,``Permutations'' datasets. The proposed dataset also performs better than other algorithms on ``Normal'' and ``Noisy Normal'' data as seen from Figure \ref{barNormal}. It works better than other algorithms on datasets containing two overlapping clusters as seen in rows ``Overlap1'' and ``Overlap2'' of  Table \ref{accures} and also in Figure \ref{barOverlap}.

Results for simulated datasets have been obtained using following parameters: The values $\mathtt{Sim2Seed}=0.6$ are used for first two datasets (``Non-linear 1'' and ``Non-linear 2'') and $\mathtt{Sim2Seed}=0.8$ for all other datasets. For all the datasets the values $\mathtt{ReuseAllSeeds} =\mathtt{FALSE}$, $\mathtt{ReuseSeedSim}=0.5$, $\mathtt{MinSeedSize}=100$, $\mathtt{ClusSim}=1$ and $\mathtt{ObsInMinBase}=3$ are used. Transformation named $norm_2()$ (mentioned in Section \ref{prePro}) is applied to datasets ``Normal'' and ``Noisy normal'', while transformation $norm_1()$ (mentioned in Section \ref{prePro}) is applied to all other datasets. Large dataset method has been used for all simulated datasets, as they have thousand or more observations.

The results for simulated datasets suggest some properties of the proposed method, which are discussed in following section. 

\subsubsection{A discussion on properties of RelDenClu as seen from experiments on Simulated datasets}
\label{prpAlgo}
It is already seen that the proposed algorithm can find relation-based biclusters with better accuracy as compared to other algorithms for the datasets as it is seen from Table \ref{accures} and Figures \ref{barNL}-\ref{barOverlap}. In this section, we discuss some of the observed properties of the proposed algorithm.

The proposed algorithm is invariant to scaling, translation and linear transforms because neither bin length nor estimated densities are affected by these transforms. Theoretically, the procedure is not invariant to arbitrary order-preserving transforms, which may change the density of the observations. We compare the accuracy for ``Base'' dataset and its various transforms by applying two-tailed $\textit{t-test}$. We find that there is no significant difference ($\textit{p-value}$ threshold is 0.01) in performance of RelDenClu for ``Scaled'', ``Translated'', ``Linear Transform'',``Square'', ``Exponential'', ``Point Proportion'' and ``Permutations'' datasets. As compared to ``Base'' dataset the performance has improved for ``Cluster Proportion'' dataset because repetition of observations in the bicluster increases the probability of its identification as high density region. The accuracies for ``Noisy uniform'' and ``Noisy Normal'' datasets are significantly lower than ``Base'' and ``Normal'' datasets, respectively. However, as compared to other algorithms RelDenClu still has better accuracy.

\subsection{Comparisons on real-life datasets}
\label{realdatares}
This section presents the performance of the proposed algorithm and other algorithms used for comparison, on three datasets obtained from the UCI ML repository by \cite{UCIML2019}. The results for datasets named Magic, Ionosphere and Breast Cancer are reported in Tables \ref{magicTab}, \ref{cancerTab} and \ref{ionTab}, respectively. 

In this section we also present an application to Supervised learning, where bicluster membership values have been used as new features to improve the performance of Naive Bayes classifier for three UCI ML datasets named Credit Card, Image Segmentation and Bioconcentration. The results are reported in Table \ref{tabresSup}.
\subsubsection{Comparison as an Unsupervised learning method}
\label{UnsupRes}

\begin{table}
	\caption{Results for Magic dataset}
	\label{magicTab}
	\begin{center}
		\begin{small}
			\begin{tabular}{|c|c|c|c|}
				\hline
				
				Method&Index Name& \multicolumn{2}{c|}{Index Value} \\
				\hline
				\multirow{4}{*}{Proposed}&Accuracy& \multicolumn{2}{c|}{\textbf{0.7374}}\\
				\cline{2-4}
				&Precision&0.87&0.59\\
				&Recall&0.70&0.81\\
				&G-score&0.78&0.69\\
				\hline
				\multirow{4}{*}{CBSC}&Accuracy& \multicolumn{2}{c|}{0.7362}\\
				\cline{2-4}
				&Precision&0.85&0.59\\
				&Recall&0.71&0.78\\
				&G-score&0.78&0.68\\
				\hline
				\multirow{4}{*}{UniBic}&Accuracy& \multicolumn{2}{c|}{0.6483}\\
				\cline{2-4}
				&Precision&0.65&*\\
				&Recall&1&0\\
				&G-score&0.81&*\\
				\hline
				\multirow{4}{*}{P3C}&Accuracy& \multicolumn{2}{c|}{0.7052}\\
				\cline{2-4}
				&Precision&0.69&0.92\\
				&Recall&0.99&0.17\\
				&G-score&0.82&0.40\\
				\hline
				\multirow{4}{*}{Subclu}&Accuracy& \multicolumn{2}{c|}{0.6104}\\
				\cline{2-4}
				&Precision&0.66&0.39\\
				&Recall&0.84&0.18\\
				&G-score&0.74&0.26\\
				\hline
				\multirow{4}{*}{ITL}&Accuracy& \multicolumn{2}{c|}{0.6154}\\
				\cline{2-4}
				&Precision&0.47&0.72\\
				&Recall&0.56&0.65\\
				&G-score&0.51&0.68\\
				\hline
				\multirow{4}{*}{Proclus}&Accuracy& \multicolumn{2}{c|}{0.6261}\\
				\cline{2-4}
				&Precision&0.66&0.41\\
				&Recall&0.89&0.14\\
				&G-score&0.76&0.24\\
				\hline
				\multirow{4}{*}{CLIQUE}&Accuracy& \multicolumn{2}{c|}{0.6845}\\
				\cline{2-4}
				&Precision&0.70&0.60\\
				&Recall&0.88&0.30\\
				&G-score&0.79&0.43\\
				\hline

			\end{tabular}
		\end{small}
	\end{center}
\end{table}

\begin{table}
	\caption{Results for Breast Cancer dataset}
	\label{cancerTab}
	\begin{center}
		\begin{small}
			\begin{tabular}{|c|c|c|c|}
				\hline
				
				Method&Index Name& \multicolumn{2}{c|}{Index Value} \\
				\hline
				\multirow{4}{*}{Proposed}&Accuracy& \multicolumn{2}{c|}{\textbf{0.9561}}\\
				\cline{2-4}
				&Precision&0.97&0.93\\
				&Recall&0.96&0.95\\
				&G-score&0.97&0.94\\
				\hline
				\multirow{4}{*}{CBSC }&Accuracy& \multicolumn{2}{c|}{0.9444}\\
				\cline{2-4}
				&Precision&0.92&1.00\\
				&Recall&1.00&0.84\\
				&G-score&0.96&0.92\\
				\hline
				\multirow{4}{*}{UniBic}&Accuracy& \multicolumn{2}{c|}{0.6501}\\
				\cline{2-4}
				&Precision&0.65&*\\
				&Recall&1&0\\
				&G-score&0.81&*\\
				\hline
				\multirow{4}{*}{P3C}&Accuracy& \multicolumn{2}{c|}{0.7965}\\
				\cline{2-4}
				&Precision&0.76&0.98\\
				&Recall&0.99&0.43\\
				&G-score&0.87&0.65\\
				\hline	
				\multirow{4}{*}{Subclu}&Accuracy& \multicolumn{2}{c|}{0.8052}\\
				\cline{2-4}
				&Precision&0.78&0.90\\
				&Recall&0.97&0.50\\
				&G-score&0.87&0.67\\
				\hline
				\multirow{4}{*}{ITL}&Accuracy& \multicolumn{2}{c|}{0.7013}\\
				\cline{2-4}
				&Precision&0.62&0.72\\
				&Recall&0.37&0.88\\
				&G-score&0.48&0.80\\
				\hline
				\multirow{4}{*}{Proclus}&Accuracy& \multicolumn{2}{c|}{0.8023}\\
				\cline{2-4}
				&Precision&0.78&0.93\\
				&Recall&0.98&0.47\\
				&G-score&0.87&0.66\\
				\hline
				\multirow{4}{*}{CLIQUE}&Accuracy& \multicolumn{2}{c|}{0.9209}\\
				\cline{2-4}
				&Precision&0.94&0.88\\
				&Recall&0.93&0.90\\
				&G-score&0.94&0.89\\
				\hline

			\end{tabular}
		\end{small}
	\end{center}
\end{table}

\begin{table}
	\caption{Results for Ionosphere dataset}
		\label{ionTab}
	\begin{center}
		\begin{small}
			\begin{tabular}{|c|c|c|c|}
				\hline
				
				Method&Index Name& \multicolumn{2}{c|}{Index Value} \\
				\hline
				\multirow{4}{*}{Proposed}&Accuracy& \multicolumn{2}{c|}{\textbf{0.9174}}\\
				\cline{2-4}
				&Precision&0.92&0.91\\
				&Recall&0.95&0.86\\
				&G-score&0.94&0.88\\
				\hline
				\multirow{4}{*}{CBSC}&Accuracy& \multicolumn{2}{c|}{0.8519}\\
				\cline{2-4}
				&Precision&0.89&0.78\\
				&Recall&0.87&0.82\\
				&G-score&0.88&0.80\\
				\hline
				\multirow{4}{*}{UniBic}&Accuracy& \multicolumn{2}{c|}{0.6410}\\
				\cline{2-4}
				&Precision&0.64&*\\
				&Recall&1&0\\
				&G-score&0.80&*\\
				\hline
				\multirow{4}{*}{P3C}&Accuracy& \multicolumn{2}{c|}{0.6000}\\
				\cline{2-4}
				&Precision&0.66&0.76\\
				&Recall&0.99&0.08\\
				&G-score&0.79&0.25\\
				\hline		
				\multirow{4}{*}{Subclu}&Accuracy& \multicolumn{2}{c|}{0.6923}\\
				\cline{2-4}
				&Precision&0.68&0.87\\
				&Recall&0.98&0.17\\
				&G-score&0.82&0.38\\
				\hline		
				\multirow{4}{*}{ITL}&Accuracy& \multicolumn{2}{c|}{ 0.5560}\\
				\cline{2-4}
				&Precision&0.34&0.32\\
				&Recall&0.29&0.50\\
				&G-score&0.32&0.65\\
				\hline
				\multirow{4}{*}{Proclus}&Accuracy& \multicolumn{2}{c|}{0.7721}\\
				\cline{2-4}
				&Precision&0.74&1.00\\
				&Recall&1.00&0.30\\
				&G-score&0.85&0.55\\
				\hline
				\multirow{4}{*}{CLIQUE}&Accuracy& \multicolumn{2}{c|}{0.7493}\\
				\cline{2-4}
				&Precision&0.72&1.00\\
				&Recall&0.1.00&0.30\\
				&G-score&0.85&0.55\\
				\hline

			\end{tabular}
		\end{small}
	\end{center}
\end{table}

\begin{table}
	\begin{center}

		\caption{Details of parameters used for Unsupervised learning on different real-life datasets}
		\label {tabParamReal}
		\resizebox{\linewidth}{!}{
		\begin{tabular}{|c|c|c|c|}
			\hline
			\backslashbox{Details}{Dataset} & \makebox{Magic}& \makebox{Cancer}& \makebox{Ionosphere}\\
			\hline
			Size of dataset&$19020 \times 10$&$683 \times 9$&$351 \times 34$\\
			Normalizing function & $norm_1()$&$norm_1()$&$norm_2()$\\
			$\mathtt{Sim2Seed}$&0.6&0.6&0.6\\
			$\mathtt{ReuseAllSeeds}$&$\mathtt{TRUE}$&$\mathtt{TRUE}$&$\mathtt{TRUE}$\\
			$\mathtt{ObsInMinBase}$ & 5& 3& 15\\
			$\mathtt{MinSeedSize}=100$& 500& 100 & 100\\
			$\mathtt{ClusSim}$& 1& 1& 1\\
			
			\hline											
		\end{tabular}
	}
	\end{center}
\end{table}

%
%
\begin{figure}[!htbp]
	\caption{Accuracy of various biclustering algorithms for unsupervised learning (calculated using Equation \ref{eqClumatch})}
	\label{barClus}
	\includegraphics[width=\linewidth]{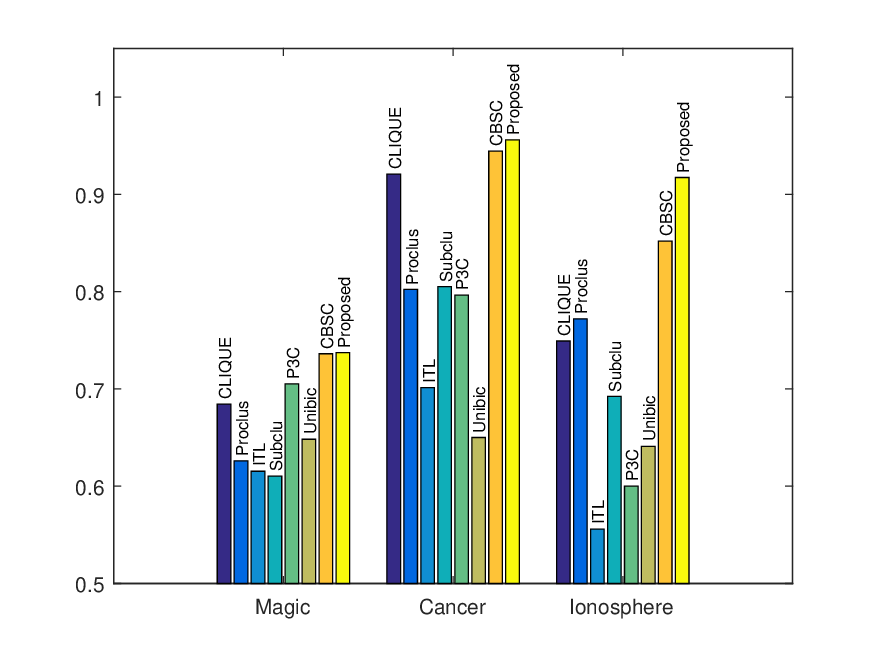}
\end{figure}

The parameter values used for real-life datasets are given in Table \ref{tabParamReal}. Since $\mathtt{ReuseAllSeeds}$ is set to $\mathtt{TRUE}$, the value of $\mathtt{ReuseSeedSim}$ is not important. For Magic dataset, we use the large dataset method to find related dense sets while for Ionosphere and Breast Cancer datasets results are reported using the small dataset procedure. For Breast Cancer dataset with 683 observations, we have also performed experiments with the large dataset method and obtained an accuracy 0.9414 and G-scores of 2 classes 0.95 and 0.99. Thus, for datasets of this size, either method gives good results.

To see whether bicluster detected by the proposed algorithm corresponds to the meaningful structure in data, we check if one of the biclusters detected corresponds to one of the known classes in data or not. Each of the datasets used contains two classes. For each observation, the membership value corresponding to a bicluster is 1 if it is included in the said bicluster and 0, otherwise. Since in each dataset observations belong to two classes, the class label can also be named as 0 and 1. For each bicluster, the number of matches between bicluster membership and the class label is calculated as below:

\begin{equation}
\label{eqClumatch}
\resizebox{0.98\linewidth}{!}{$Accuracy=  \frac{max(sum(XNOR(Bi\_mem(O),Class(O))), sum(XOR(Bi\_mem(O),Class(O)))}{length(O)}$}
\end{equation}

where $Bi\_mem$ denotes the bicluster membership, $O$ denotes the set of observations in entire dataset and $Class$ denotes the class label. $XNOR$ and $XOR$ are the logical operators and $length()$ denotes the size of $O$. For the bicluster obtaining the best accuracy according to equation \ref{eqClumatch}, precision, recall and G-score are also reported for both the classes. It is seen that the proposed method yields the best accuracy for all the three datasets used for investigation. Though other methods attain better precision or recall in some cases, the proposed method attains better G-score in all the cases. For UniBic it is seen that best accuracy is obtained when all the observations are assigned to a bicluster, therefore precision and recall is reported only for one class and the other column is marked with an asterix (*). For better visualization, accuracies obtained by different methods are presented in Figure \ref{barClus} as bar graphs. 

\subsubsection{Usefulness of the proposed algorithm to aid supervised learning}
\label{supRes}

\begin{table}
	\begin{centering}

		\caption{Details of parameters used for Classification on different real-life datasets}
		\label {tabParamRealSup}
		\resizebox{\linewidth}{!}{
			\begin{tabular}{|c|c|c|c|}
				\hline
				\backslashbox{Details}{Dataset} & \makebox{Credit card}& \makebox{Image Segmentation}& \makebox{Bioconcentration}\\
				\hline
				Size of dataset&$30000 \times 24$&$2310 \times 19$&$779 \times 14$\\
				Normalizing function & $norm_1()$&$norm_1()$&$norm_1()$\\
				$\mathtt{Sim2Seed}$&0.6&0.6&0.9\\
				$\mathtt{ReuseAllSeeds}$&$\mathtt{FALSE}$&$\mathtt{FALSE}$&$\mathtt{TRUE}$\\
				$\mathtt{ReuseSeedSim}$& 0.9& 0.8& NA \\
				$\mathtt{ObsInMinBase}$ & 7& 7& 7\\
				$\mathtt{MinSeedSize}=100$& 1000& 100 & 100\\
				$\mathtt{ClusSim}$& 1& 0.5& 0.6\\
				
				\hline											
			\end{tabular}
		}
	\end{centering}
\end{table}

\begin{table*}[!htbp]
	\caption{Classification accuracy (after adding features generated using diffrent biclustering methods) }
	\label{tabresSup}
	\resizebox{\textwidth}{!}{
		\begin{tabular}{|c|c|c|c|c|c|c|c|c|c|c|}
			\hline
			\multirow{2}{*}{Dataset}&\multirow{2}{*}{Accuracy}	&\multicolumn{9}{c|}{Methods used}\\
			\cline{3-11}
			
			&&Original&CLIQUE&Proclus&ITL&Subclu&P3C&UniBic&CBSC&Proposed 	\\			
			
			\hline
			\multirow{2}{*}{Credit card} &Mean&0.696&0.551&0.722&0.659&0.689&0.702&0.679&0.670&\textbf{0.766}\\
			&Deviation&0.028&0.006&0.016&0.024&0.015&0.030&0.011&0.021&0.004\\
			\hline
			\multirow{2}{*}{Image Segmentation}&Mean&0.795&0.818&0.735&0.796&0.794&0.761&0.740&0.693&\textbf{0.860}\\
			&Deviation&0.015&0.030&0.013&0.023&0.024&0.022&0.030&0.023&0.016\\
			
			\hline
			\multirow{2}{*}{Bioconcentration} &Mean&0.601&0.577&0.611&0.605&0.606&0.578&0.580&0.606&\textbf{0.818}\\
			&Deviation&0.048&0.036&0.063&0.040&0.057&0.045&0.059&0.061&0.046\\
			\hline
		\end{tabular}
	}
\end{table*}

\begin{figure}
	\caption{Improvement in classification accuracy using various biclustering methods }
	\label{barSup}
	\includegraphics[width=\linewidth]{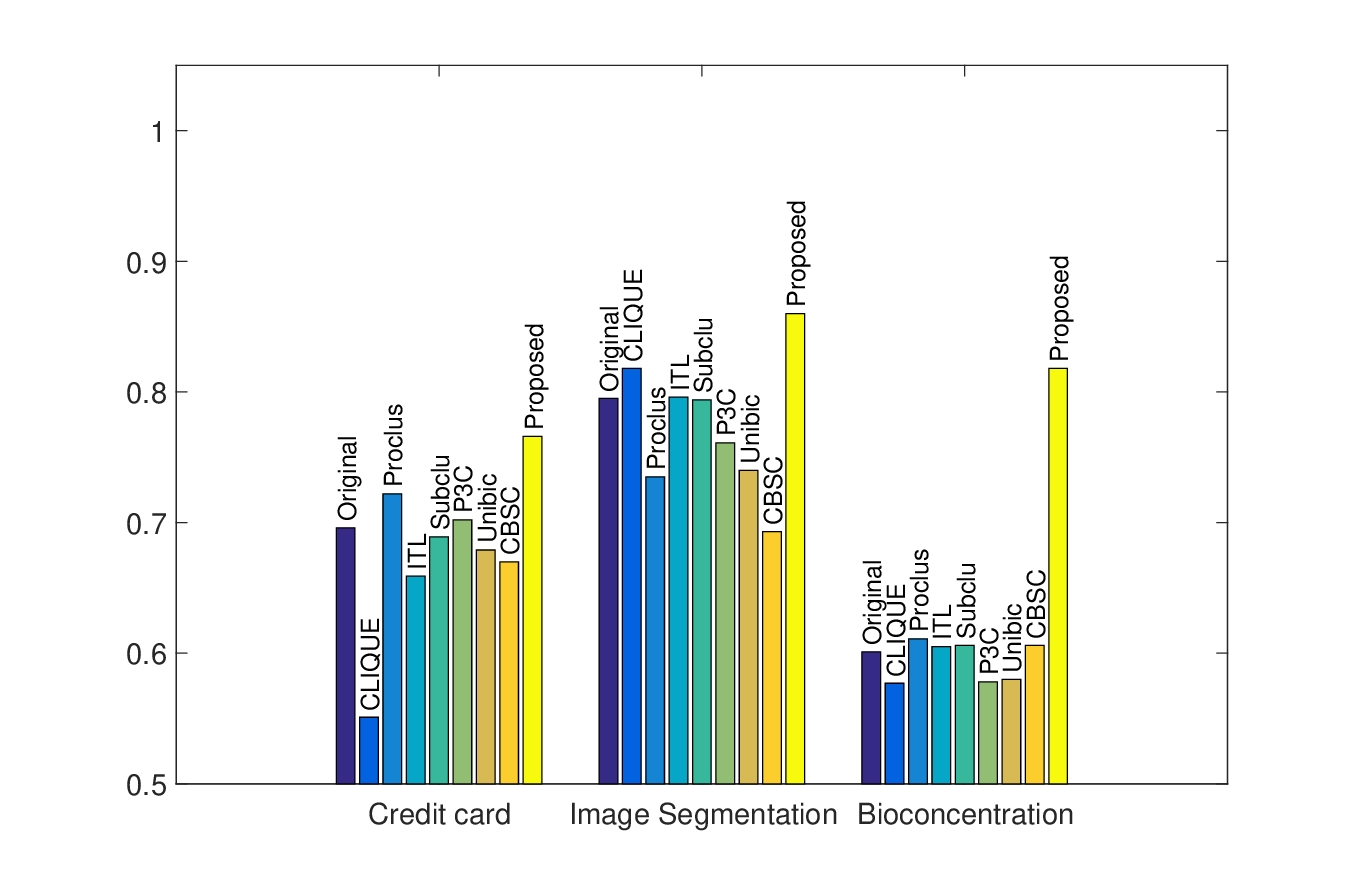}
\end{figure}

As another application, we have used RelDenClu and other biclustering methods to generate new features that are further used to improve classification performance on three UCI-ML datasets (described in Section \ref{supRealds}). We can see that these datasets belong to diverse fields and finding solutions to classification problems associated with them has practical implications. To demonstrate the effectiveness of the proposed algorithm for real-life datasets, the results are compared with those obtained using seven other state-of-the-art techniques as used in earlier investigations. We try to improve the classification accuracy for the datasets by adding new binary features generated using biclustering algorithms.

For each dataset, biclusters are found using different methods. For a given algorithm, each bicluster obtained results in a new feature. The an observation belongs to this bicluster the feature value is taken as 1 otherwise 0, i.e. the corresponding membership value is taken as a new feature. Naive Bayes classifier is used after adding new features to a database.

We find that the features generated by RelDenclu provide a greater improvement in accuracy as compared to other algorithms for the three datasets. The large dataset method has been used for these datasets. The parameters used have been reported in Table \ref{tabParamRealSup} and the results are reported in Table \ref{tabresSup}. The accuracies obtained using Naive Bayes classifier for original and enhanced datasets are depicted in Figure \ref{barSup}.
\subsection{Execution time and performance on large datasets}
\label{relexec}
 \begin{table*}[!htbp]
 	
 	\caption{Execution time (in seconds) for datasets corresponding to Table \ref{accures}}
 	\label{runtimetab}
 	\resizebox{\textwidth}{!}{
 		\begin{tabular}{|c|c|c|c|c|c|c|c|c|}
 			
 			\hline
 			
 			Dataset&CLIQUE&Proclus&ITL&Subclu&P3C&UniBic&CBSC&Proposed\\
 			\hline
 			
 			\hline
 			Non-Linear 1 &0.0105&0.0595&9.0962&0.0642&0.3587&0.583&37.464&2.7663\\
 			\hline
 			Non-Linear 2&0.0065&0.0545&9.2478&0.0515&0.6502&0.612&31.82&1.0136\\
 			\hline
 			Base &0.105&0.052&3.171&0.056&0.513&0.583&42.416&5.204\\
 			\hline
 			Scaled&0.015&0.055&3.302&0.043&0.955&0.612&51.4&5.25\\
 			\hline
 			Translated&0.015&0.049&3.498&0.05&0.952&0.649&43.964&5.3\\
 			\hline
 			Linear transform&0.018&0.055&3.276&0.055&0.433&0.673&41.313&5.532\\
 			\hline
 			Square&0.04&0.057&3.544&0.048&3.237&0.56&37.105&2.733\\
 			\hline
 			Exp&0.02&0.051&3.485&0.054&0.224&0.573&43.985&2.511\\
 			\hline
 			Point proportion&0.033&0.12&9.994&0.12&1.083&2.596&91.973&36.437\\
 			\hline
 			Cluster proportion&0.021&0.088&6.519&0.082&0.848&1.436&83.796&29.625\\
 			\hline
 			Noisy uniform&0.014&0.056&2.94&0.049&0.458&0.621&42.304&13.312\\
 			\hline
 			Permutations&0.013&0.063&3.272&0.07&0.588&0.583&41.654&5.468\\
 			\hline
 			Normal &0.071&0.054&1.187&0.071&18.84&0.531&66.659&15.896\\
 			\hline
 			Normal noisy&0.044&0.052&1.259&0.061&9.75&0.577&63.821&18.27\\
 			\hline
 			Overlap cluster &1.041&0.061&2.089&0.06&9.36&0.608&113.15&47.19\\

 			\hline
 		\end{tabular}
 	}
 \end{table*}

 \begin{table*}[!htbp]

 	\caption{Accuracy and execution time (in seconds) for dataset of size $20000 \times 100$ having a bicluster of size $10000 \times 30$}
 	\label{exectimelarge}
 	\resizebox{\textwidth}{!}{
 		\begin{tabular}{|c|c|c|c|c|c|c|c|}
 			
 			\hline
 			Method&CLIQUE&Proclus&ITL&Subclu&UniBic&CBSC&Proposed\\
 			\hline
 			Accuracy &0.868&0.862&0.612&0.861&0.735&0.903&\textbf{0.9796}\\
 			\hline
 			Execution time&3.4&22.814&10106&19.17&678.972&12066.662&1396.400\\

 			\hline
 		\end{tabular}
 	}
 \end{table*}

Execution time is considered as another evaluation criterion and the results are reported in Table \ref{runtimetab}. The datasets used here are the same as those used in Table \ref{accures}. As seen from Table \ref{runtimetab}, the execution time for the proposed method is high compared to several density-based methods like Subclu, Proclus, CLIQUE. However, its performance is significantly better as seen in Table \ref{accures}. The algorithm is executed with Intel Core i7 CPU and 8 GB memory.

Table \ref{exectimelarge} reports execution times and accuracy for a large dataset of size $20000 \times 100$. The data used, has been generated in a similar way as ``Base'' dataset details are given in Appendix \ref{artDatDet}. The parameters used by RelDenClu are $\mathtt{Sim2Seed}=0.95$, $\mathtt{ReuseAllSeeds} =\mathtt{FALSE}$, $\mathtt{ReuseSeedSim}=0.5$, $\mathtt{MinSeedSize}=5000$, $\mathtt{ClusSim}=1$ and $\mathtt{ObsInMinBase}=15$. Transformation named $norm_1()$ (mentioned in Section \ref{prePro}) is applied. In this table, execution time of P3C is not reported as R package for this method could not execute the algorithm for the given dataset, on given system configuration. It is noticed that, the proposed algorithm has a lower execution time as compared to CBSC.

Time complexity of ITL has been reported by \cite{ITL2003}, while time complexity of other methods used for comparison has been reported by \cite{CBSC2018}. We are reporting the time complexity of the proposed method, where $N$ and $M$ respectively, are the number of observations and features in a given dataset and $m_{total}=m_1+m_2+, \cdots + m_k$ where $m_i$ gives the number of features in the $i^{th}$ bicluster with a total of $k$ biclusters present. Note that $m_{total}$ corresponds to the total number of columns in the complete set of biclusters i.e., before removing similar biclusters using cosine measure, and without leaving out seed biclusters which are similar. The proposed method has time complexity $N^2M+NM^3+Nm_{total}^2$ and $NM+NM^3+Nm_{total}^2$ for small and large datasets, respectively. Theoretically, the time complexity of CBSC and the proposed method are similar. However, the execution time for the proposed method is found to be much smaller as compared to CBSC for the high dimensional dataset. This happens because the proposed method does not calculate Minimal Spanning tree for each pair of dimensions, which are needed in CBSC.
  

\section{Applicability of proposed method for COVID-19}
\label{covidAn}

\begin{table*}[!htbp]
	\begin{centering}

		\caption{Features (from WDI dataset) lying in bicluster containing countries having high COVID-19 occurrence (90 percentile ) in January 2020}
		\label {tabFeatCovid}
		\resizebox{\textwidth}{!}{
			\begin{tabular}{|c|m{0.46\textwidth}|m{0.24\textwidth}|m{0.075\textwidth}|m{0.075\textwidth}|m{0.075\textwidth}|m{0.075\textwidth}|}
				\hline
				S.No.&Feature description&Feature Indicator&$\rho $(Jan)&$\rho_s $(Jan)&$\rho $(Dec)&$\rho_s $(Dec)\\
				\hline
				
		1&Cause of death, by communicable diseases and maternal, prenatal and nutrition conditions (\% of total)&SH.DTH.COMM.ZS&-0.023 &-0.045&0.5376&0.6775\\ \hline 
		2&Cause of death, by non-communicable diseases (\% of total)&SH.DTH.NCOM.ZS&0.037  &0.068&0.56&0.664\\ \hline
		3&GDP per capita, PPP (current international \$)&NY.GDP.PCAP.PP.CD&0.285  &0.159 &0.4435&0.6175\\ \hline
		4&Incidence of tuberculosis (per 100,000 people)&SH.TBS.INCD&-0.044  &0.003&0.386&0.5039\\ \hline
		5&Life expectancy at birth, total (years)&SP.DYN.LE00.IN&0.142  &0.187&0.5466	&0.6401\\ \hline
		6&Mortality rate, adult, female (per 1,000 female adults)&SP.DYN.AMRT.FE&-0.113 &-0.195&0.5197	&0.6623\\ \hline
		7&Mortality rate, adult, male (per 1,000 male adults)&SP.DYN.AMRT.MA&-0.141 &-0.159&0.453	&0.5659\\ \hline
		8&Population ages 15-64 (\% of total)&SP.POP.1564.TO.ZS&0.157  &0.085&0.4295	&0.4616\\ \hline
		9&Population ages 65 and above (\% of total)&SP.POP.65UP.TO.ZS&0.054  &0.174&0.546	&0.5881\\ \hline
		10&Population density (people per sq. km of land area)&EN.POP.DNST&0.347  &0.238&0.0258	&0.1607\\ \hline
		11&Population, total&SP.POP.TOTL&-0.014  &0.334&0.0924	&0.1608\\ \hline
		12&Survival to age 65, female (\% of cohort)&SP.DYN.TO65.FE.ZS&0.121  &0.195&0.5457&	0.6641\\ \hline
		13&Survival to age 65, male (\% of cohort)&SP.DYN.TO65.MA.ZS&0.149  &0.168&0.4695	&0.574\\ \hline
		14&Trade (\% of GDP)&NE.TRD.GNFS.ZS&0.363 &0.000&0.3117	&0.2618\\ \hline

			\end{tabular}
		}
	\end{centering}
\end{table*}

\begin{figure*}[]
	\caption{Number of people (per million) suffering from COVID-19 on 31 January 2020 vs relevant features obtained using proposed algorithm. The countries which lie in above ninty percentile are marked with blue circle.}
	\label{coviFimg}
	\includegraphics[width=\textwidth]{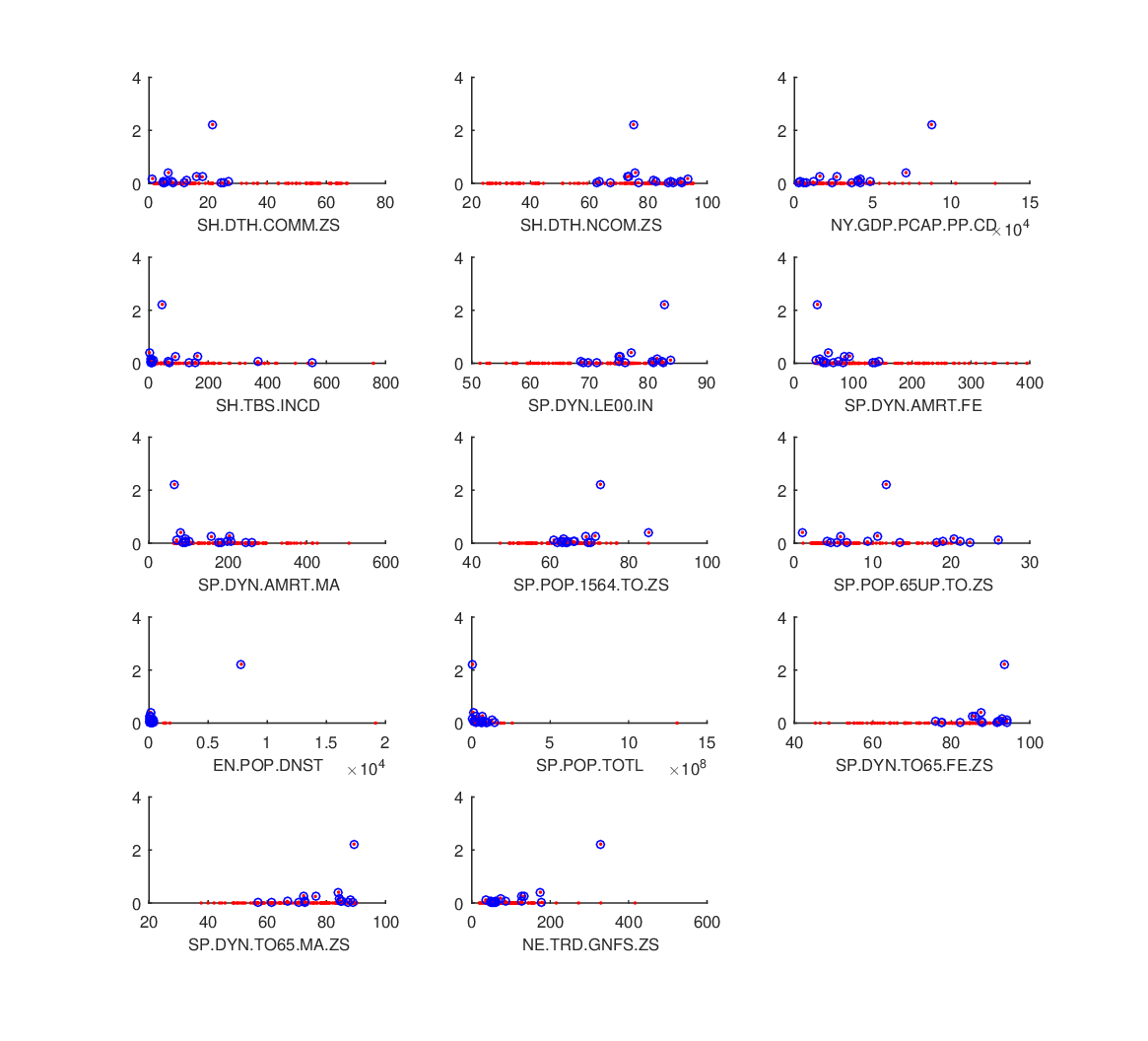}
\end{figure*}
Toward the end of 2019, a disease named COVID-19 emerged as a highly contagious disease and has resulted in a pandemic of unprecedented scale. However, it is seen that some countries are affected by this disease to a greater extent than others. During the early phase of spread of COVID-19, policy makers and medical practitioners were faced with a lack of information regarding factors affecting the spread of the disease. 

Using the data for month of January 2020, we found that the proposed method identified some factors related to disease occurrence rate. At this time, the Pearson and Spearman correlations of selected features with disease occurrence were low. However, the Pearson and Spearman correlations of the selected features with disease occurrence on 31 December 2020 are high, indicating that the selected features are indeed important. During the early phases of the disease these relations were being obscured by data from countries where the disease was still in nascent stages. The proposed method is capable of overcoming this problem as it searches for relations between subsets of data. An early detection of factors impacting the spread of diseases makes the proposed method a useful tool for understanding new diseases and epidemics. The details of the methodology and results are provided in following sections.



\subsection{Use of RelDenClu for finding relevant features in COVID-19 dataset}
For this analysis the data has been obtained from two sources. The first one called the WDI dataset henceforth, is taken from the site of \cite{WBwdi}. Twenty-seven attributes that intuitively seem relevant to the spread of COVID-19 and do not contain too many missing values were selected, for further analysis. The list of World Development indicators in this dataset is provided in the appendix \ref{apptable}. The second dataset contains the cumulative number of COVID-19 cases detected in a country on each day, and the same has been provided by \cite{Dong2020}. The parameters used by RelDenClu are  $\mathtt{Sim2Seed}=0.98$, $\mathtt{ReuseAllSeeds}=\mathtt{FALSE}$, $\mathtt{ReuseSeedSim}=0.7$, $\mathtt{ObsInMinBase}=10$ and $\mathtt{MinSeedSize}=15$, $\mathtt{ClusSim}=0.75$. The normalization used is $norm_1()$. 

To understand the features affecting the infection rate, we use the number of confirmed COVID-19 cases for the month of January 2020 i.e., the cumulative number of confirmed cases each day of January. This along with WDI features is used as input for RelDenClu. We define the infection rate as the number of confirmed COVID-19 cases per million people living in a region. We find countries lying above 90 percentile in terms of infection rate. The bicluster having a maximum match with this set is identified among all the biclusters obtained using RelDenClu. For a given country we say that there is a match between a bicluster $B$ and 90 percentile set if the country belongs to both or does not belong to either. Since the biclusters are obtained based on the similarity between features for chosen observations, it can be inferred that the relationship between these features distinguishes the given set of observations from other observations. Thus these features are likely to affect the spread of the disease.

\subsection{Relevant features in COVID-19 dataset obtained using RelDenClu during early stages of the pandemic}

The proposed method selected 14 out of 27 WDI features that are capable of distinguishing regions with high infection rate. Table \ref{tabFeatCovid} lists these 14 features (Feature Code and Feature Description) obtained by RelDenClu using the data from January 2020 and also provides the values of the Correlation coefficient ($\rho$) and Spearman correlation ($\rho_s$) between the chosen feature and the infection rate of COVID-19 for various countries for 31 January 2020 and 31 December 2020. From this table we find that 10 out of 14 features selected from biclusters using the disease occurrence rate have a Spearman correlation greater than 0.5 with the disease occurrence rate on 31 December 2020. However, Reldenclu could identify the importance of these features much earlier i.e. from the January data when the Spearman correlation could not reflect this relation. Amongst the features not included in the bicluster, only 4 out of the 13 features have a Spearman correlation greater than 0.5 with disease occurrence rate on 31 December 2020. This suggests that the proposed algorithm could be explored to identify important factors impacting a new disease and, this, in turn,  will aid in preparing a strategy, in a more scientific manner, to combat the pandemic situation across countries. Indeed some of these selected features for COVID-19 have already been studied and are found to be important, as discussed in the following paragraphs, where we refer to features using their serial number (S.No.) given in Table \ref{tabFeatCovid}. 

Among the features selected by RelDenClu, the percentage of deaths in region due to communicable and non-communicable diseases (S. no. 1 and 2 of Table \ref{tabFeatCovid}), falls in line with the report given by \cite{CDCcovRep}. We find that age-related features (S. no. 8 and 9 of Table \ref{tabFeatCovid}) are also selected. Features showing life expectancy and mortality rate for overall population and for each gender also affect the age distribution of people in a region (S. no. 5, 6, 7 of Table \ref{tabFeatCovid} ) are selected by RelDenClu. The feature indicating the incidence of tuberculosis (S. no. 4 of Table \ref{tabFeatCovid} ), is interesting. It is also seen that countries with higher tuberculosis incidences have relativity lower COVID-19 infection rates. A study by \cite{Curtis2020} has already noted that administering BCG vaccine is likely to reduce severity of COVID-19 symptoms. Another factor that may contribute to this association is that tuberculosis is more rampant in warm humid climates while COVID-19 is expected to spread faster in cold dry climate as noted by \cite{NScovClimate}.	

The feature showing  per capita GDP (S. no. 3 in Table \ref{tabFeatCovid}) may be related to higher COVID-19 rate due to higher number of tests. The selected features showing total population and population density as well as trade as percentage of GDP (S. no. 10, 11 and 14 of Table \ref{tabFeatCovid}) are not strongly related to COVID-19 occurence rate but are selected by proposed method, probably due to indirect relation.


In a nutshell, most of features selected by the proposed algorithms are relevant and given methodology and allows us to explore some previously unknown associations. Once features are identified from larger datasets, a more detailed analysis could be done for individual features.

The algorithm has also been used to analyze biclusters based on COVID occurence rate for December 2020 and selected features have been reported in Table \ref{tabFeatCovidDec} presented in Appendix \ref{apptable}.



\section{Conclusion}
\label{Conc}
In this article, we proposed an algorithm that finds biclusters based on non-linear relations between features. By analysing local variations in density, the algorithm is seen to perform well on non-linear datasets. Experiments on simulated datasets have shown that the proposed algorithm is consistent under linear transforms. It is also seen to provide consistent performance under many non-linear transformations, yielding improved results as compared to seven other methods on noisy datasets. The significance of these results has been shown using $\textit{t-test}$.

The proposed algorithm is seen to be effective in discovering existing classes for three real-life datasets of UCI ML repository (Magic, Cancer, Ionosphere). Additionally, RelDenClu provides higher accuracy when considered as a precursor for supervised learning (Credit card, Image Segmentation, Bioconcentration). It has also been applied to a dataset containing information about number of COVID-19 cases in different regions and respective development indicators, to obtain factors (demographic and others) impacting number of confirmed infections in a region, and we found some interesting associations which are yet to be studied further by the experts. This information is likely to be useful to policy makers as well as medical researchers. Finally, the proposed algorithm facilitates us to understand the relationship between subsets of a dataset which is otherwise obscured by unrelated subsets of observations or overlooked due to non-linearity. 

\begin{acknowledgements}
The authors are grateful to the Science and Engineering Research Board (SERB), Department of Science and Technology (DST), Government of India for supporting this research through sanctioning a research project under MATRICS scheme (File Number: MSC/2020/000409).	

\end{acknowledgements}

\bibliographystyle{spbasic}      
\bibliography{bicluster}   
\newpage
\appendix
\FloatBarrier
\clearpage
\noindent{\textbf{Appendix}}

\vspace{2em}

\section{Detailed procedure for generating simulated datasets}
\label{artDatDet}
This section outlines our method of generating datasets for this purpose. The datasets for the first and the second rows require several functions which are reported in Table \ref{tabfuncnl1}. Rest of the data is generated using a single function.

\begin{table}
	\begin{center}

		\caption{Functions used to generate datasets of type Non-Linear 1 and Non-Linear 2.Both the datasets are similar but range of some features is different.}
		\label {tabfuncnl1}    
		
		\begin{tabular}{|c|r|r|}
			\hline
			$h_i$ & Non-Linear 1& Non-Linear 2\\
			\hline
			$h_1(x)$& $I(x)=x$& $I(x)=x$\\
			$h_2(x)$& $\sin(x)$ & $\sin(x)$\\
			$h_3(x)$& $x^2$& $x^2$\\
			$h_4(x)$& $x^{10}$& $x^{10}$\\
			$h_5(x)$& $\sin(\pi x)$& $0.5\sin(\pi x)$\\
			$h_6(x)$& $\sin(2 \pi x)$& $0.5\sin(2 \pi x)+0.5$\\   
			$h_7(x)$& $x^3$& $x^3$\\    
			$h_8(x)$& $4x^2$& $x^2$\\
			$h_9(x)$& $\sin(4 \pi x)$& $0.5 \sin(4 \pi x) +0.5$\\
			
			$h_{10}(x)$& $4x^3$& $x^3$\\   
			\hline                                            
		\end{tabular}
		
	\end{center}
\end{table}

In Table \ref{accures}, the first row shows the results on datasets generated in the following manner. A random matrix of size $1000 \times 20$ is generated. Now we will replace $500 \times 10$ submatrix of data in a way so that the $i^{th}$ column of this submatrix is given by $h_i(x)$, where $x$ is the value in the first column and the definitions of $h_i$ for $i=1, 2, \cdots, 10$ are given in Table \ref{tabfuncnl1}. The row and column numbers that form the bicluster submatrix are chosen randomly. Thus we embed a bicluster of size $500 \times 10$ in the data. Note that for some columns, the range in which the bicluster elements lie, is different from the range of remaining data. Thus the background for the biclusters is highly variable. One such example was presented in Figure \ref{figVarNM} in Section \ref{toyex}.

Datasets used for results shown in the second row of Table \ref{accures} are generated using a set of functions, which are modifications of the functions used in the first row. The modifications are made so that data in bicluster lies in the same range as the rest of the data. The modified functions have been reported in the second column of Table \ref{tabfuncnl1}. It may be noted that the function $x^2$ occurs both as $h_3$ and $h_8$ in the second column and $x^3$ occurs both as $h_7$ and $h_{10}$, amounting to the repetition of a column in the bicluster. However, we have retained the repeated columns to retain the similar structure of bicluster as in datasets for the first row.

The third row presents the results on datasets of size $1000 \times 20$ drawn from uniform distribution. It has a bicluster of size $500 \times 10$ generated using functions $h_1(x) = I(x) = x$, where I is the identity function and $h_i(x)= a_i*x$ with $i=2, 3 \cdots, 10$. $a_i$ with $i=2, 3, \cdots,10$ are different random values lying in interval $(0,1)$. This is our base data for Uniform distribution datasets. Different transformations have been applied to this data for analysing properties of biclustering algorithms and corresponding results have been reported in following rows.

The fourth row contains results for datasets obtained by scaling each column of data generated for the third row with a random number lying in the interval $(0,1)$. The fifth row contains results for datasets obtained by adding a random number lying in $(0,1)$ to each column of the data generated for the third row. These are used to analyze the scaling and translation properties of the algorithms.
The sixth row contains results for datasets obtained by linear transform to each column of data generated for the third row. In a way, this is a combination of scaling and translation. Two random numbers $r_1$ and $r_2$ are generated for each column and the transform $r_1 x + r_2$ is applied. Scaling, translation and linear transforms are special cases of distance preserving transforms.

The seventh row and the eighth row contain results for datasets obtained by applying square transform $f(x)=x^2$ and exponential transform $f(x)=\exp(x)$, to each observation and each feature of data generated for the third row. Since data used here lies in the range $(0,1)$ both of these are monotone transforms, but not necessarily distance preserving.

The ninth row contains results for datasets obtained by duplicating each observation of datasets used for reporting the results in the third row. Thus these datasets contain 2000 observations.
The tenth row contains results for datasets obtained by duplicating each observation of bicluster in datasets generated for the third row. Thus these datasets contain 1500 observations.
The eleventh row contains results for datasets obtained by adding a random number in range (0,0.1) to each element of data matrices. The random noise is drawn from a uniform distribution. The robustness of the algorithms regarding noise is checked through this.
The twelfth row contains results for datasets obtained by randomly shuffling rows and columns of the data generated for the third row.

The thirteenth row presents the results on datasets of size $1000 \times 20$ drawn from Gaussian distribution. The bicluster submatrix of size $500 \times 10$ is generated in the same way as the third row. 
The fourteenth row contains results for datasets obtained by adding noise to each element of the data matrix generated in the thirteenth row. Noise is generated using random numbers drawn from Gaussian distribution with the standard deviation $0.1$. Thus we test the robustness of the algorithms for Gaussian data.

The fifteenth and the sixteenth rows present the results on data of size $1000 \times 20$ drawn from uniform distribution. It has 2 biclusters of sizes $500 \times 10$ and 
$300 \times 8$. The fifteenth and the sixteenth rows give the results for the first and second biclusters, respectively. These biclusters are generated by adding a randomly chosen value to its elements. Thus elements of biclusters lie in a different range from the elements not lying in the bicluster. These biclusters have an overlapping area of $300 \times 3$. This is done to see if the algorithm is capable of finding overlapping biclusters.

\subsection {Generating large simulated dataset}
Initially data is generated using uniform distribution on $[0,1]\times[0,1]$. Then, bicluster of size $10000 \times 30$ using functions $h_1(x) = I(x) = x$, where $I$ is the identity function and $h_i(x)= a_i*x$ for $i=2, 3, \cdots, 30$. The term $a_i$ for $i=2, 3, \cdots,30$ are different random values lying in interval $(0,1)$. 
 
\clearpage
\section{Tables}
\label{apptable}
\noindent{Features used from WDI dataset\\
	\begin{tabular}{lp{0.6\textwidth}}
		\hline
		Air transport, passengers carried\\                                                                    
		Cause of death, by communicable diseases and maternal, prenatal and nutrition conditions (\% of total)\\
		Cause of death, by non-communicable diseases (\% of total)\\                                            
		Current health expenditure per capita, PPP (current international \$)\\                                 
		Death rate, crude (per 1,000 people)\\        
		GDP per capita, PPP (current international \$)\\                                                        
		Incidence of tuberculosis (per 100,000 people)\\                                                       
		International migrant stock, total\\                                                                   
		International tourism, number of arrivals\\                                                            
		Labor force participation rate, total (\% of total population ages 15+) (modeled ILO estimate)\\        
		Life expectancy at birth, total (years)\\                                                              
		Mortality from CVD, cancer, diabetes or CRD between exact ages 30 and 70 (\%)\\                         
		Mortality rate, adult, female (per 1,000 female adults)\\                                              
		Mortality rate, adult, male (per 1,000 male adults)\\                                                  
		Out-of-pocket expenditure (\% of current health expenditure)\\                                          
		People using at least basic sanitation services (\% of population)\\                                    
		PM2.5 air pollution, population exposed to levels exceeding WHO guideline value (\% of total)\\         
		Population ages 15-64 (\% of total)\\                                                                   
		Population ages 65 and above (\% of total)\\                                                            
		Population density (people per sq. km of land area)\\                                                  
		Population, total\\                                                                                    
		Survival to age 65, female (\% of cohort)\\                                                             
		Survival to age 65, male (\% of cohort)\\                                                               
		Trade (\% of GDP)\\                                                                                     
		Tuberculosis case detection rate (\%, all forms)\\                                                      
		Tuberculosis treatment success rate (\% of new cases)\\                                                 
		Urban population (\% of total)\\ 
		
	\end{tabular}

	\clearpage

	\noindent{Symbols used in this article}
	
	\begin{tabular}{cp{0.6\textwidth}}
		\hline
		$i,j,k,p,q,oi,fi $ & Used as indices for rows, columns, biclusters, cells of grids etc. \\
		$D$ & Data matrix\\
		$O$ & Set of observations\\
		$F$ & Set of features\\
		$A$ & Submatrix of D \\
		$A_{*,j}$ &$j_{th}$ column of matrix $A$\\
		$A_{i,*}$ &$i_{th}$ row of matrix $A$\\
		$A_{i,j}$ &The element in $i_{th}$ row and $j_{th}$ column of matrix $A$\\	
		$N$ & Number of observations in $D$ \\
		$M$ & Number of features in $D$\\
		$n$ & Number of observations in a bicluster\\
		$m$ & Number of features in a bicluster\\
		$X, Y$ & Names used to refer to two axes in two-dimensional space\\
		$v_x, v_y$ & Coordinates for centre of a cell in two-dimensional space given by $X$ and $Y$\\
		$x_{len}, y_{len}$ & Size of a cell along $X$ and $Y$ axes\\
		$s$& Maximal separation along any particular axis\\
		$s_x, s_y$ & Maximal separation along $X$ and $Y$ axes\\
		$c, C_1, C_2$ & Constants\\
		$A_N$ & Area of a cell when data of size $N$ is partitioned in two dimensional space\\
		$n_X, n_Y$ & Number of partitions along $X$ and $Y$ axes\\
		$P_X(i)$ & Probability of observation lying in $i_{th}$ bin along $X$ axis\\
		$P_Y(j)$ & Probability of observation lying in $j_{th}$ bin along $Y$ axis\\
		$P_{XY}(i)$ & Probability of observation lying in $i_{th}$ bin along $X$ axis and $j_{th}$ bin along $Y$ axis\\	
		$E_{XY}[f_{XY}]$ & Expectation of a function $f_{X,Y}$ with reference to joint distribution of $X$ and $Y$\\
		$e_{i,j}$ & The element of $D$ in $i_{th}$ row and $j_{th}$ column, subscript notation has been used instead of matrix notation for better readability\\
		$x$ & Used to denote argument of function\\
		$f(x)$ & Function of argument $x$\\
		$f_1, f_2, f_3, \cdots$ & Used to denote unique features in the dataset (not necessarily consecutive)\\
		$S, S'$ & Seed biclusters found in Step \ref{remnoise}\\
		$B$& Used to denote the seed bicluster while it grows iteratively in Step \ref{getlargebi}\\
		$O_1, O_2$ & Observation sets for the first and the second bicluster\\
		$F_1, F_2$ & Feature sets for the first and the second bicluster\\	
		$\#()$& Cardinality of set\\
		$r, \epsilon$& Two parameters used by Subclu which represent the radius of disc used for density estimation and threshold for number of observations in each disc\\
		$h_1(x), h_2(x), \cdots$& Different functions of variable $x$\\
		$I(x)$& Identity function\\
		$r_1, r_2$& Two random variables used for linear transformation of data\\
		$a_1, a_2, \cdots, a_i$ & Random variables used for generating simulated data\\
		$m_B$& Membership matrix for bicluster under consideration\\
		$m_E$& Estimated membership matrix for bicluster under consideration\\
		$m_1, m_2, \cdots, m_k$ & Number of features in biclusters indexed by $1, 2, \cdots, k$\\	
		$m_{total}$& Sum of number of features in each bicluster obtained for a particular dataset\\
		
	\end{tabular}\\
	
\begin{table*}[!htbp]
	\begin{centering}

		\caption{Features (from WDI dataset) lying in bicluster containing countries having high COVID-19 occurrence (90 percentile ) in December 2020}
		\label {tabFeatCovidDec}
		\resizebox{\textwidth}{!}{
			\begin{tabular}{|c|m{0.46\textwidth}|m{0.24\textwidth}|m{0.075\textwidth}|m{0.075\textwidth}|}
				\hline
				S.No.&Feature description&Feature Indicator&$\rho $(Dec)&$\rho_s $(Dec)\\
				\hline
				
	1&Cause of death, by communicable diseases and maternal, prenatal and nutrition conditions (\% of total)&SH.DTH.COMM.ZS&-0.538 &-0.677 \\ \hline
	2&Cause of death, by non-communicable diseases (\% of total)&SH.DTH.NCOM.ZS&0.560  &0.664\\ \hline
	3&Death rate, crude (per 1,000 people)&SP.DYN.CDRT.IN&0.169  &0.070\\ \hline
	4&Incidence of tuberculosis (per 100,000 people)&SH.TBS.INCD&-0.386 &-0.504\\ \hline
	5&Life expectancy at birth, total (years)&SP.DYN.LE00.IN&0.547  &0.640\\ \hline
	6&Mortality from CVD, cancer, diabetes or CRD between exact ages 30 and 70 (\%)&SH.DYN.NCOM.ZS&-0.332 &-0.395\\ \hline
	7&Mortality rate, adult, female (per 1,000 female adults)&SP.DYN.AMRT.FE&-0.520 &-0.662\\ \hline
	8&Mortality rate, adult, male (per 1,000 male adults)&SP.DYN.AMRT.MA&-0.453 &-0.566\\ \hline
	9&Out-of-pocket expenditure (\% of current health expenditure)&SH.XPD.OOPC.CH.ZS&-0.215 &-0.249\\ \hline
	10&People using at least basic sanitation services (\% of population)&SH.STA.BASS.ZS&0.521  &0.633\\ \hline
	11&Population ages 15-64 (\% of total)&SP.POP.1564.TO.ZS&0.430  &0.462\\ \hline
	12&Population ages 65 and above (\% of total)&SP.POP.65UP.TO.ZS&0.546  &0.588\\ \hline
	13&Population, total&SP.POP.TOTL&-0.092 &-0.161\\ \hline
	14&Survival to age 65, female (\% of cohort)&SP.DYN.TO65.FE.ZS&0.546  &0.664\\ \hline
	15&Survival to age 65, male (\% of cohort)&SP.DYN.TO65.MA.ZS&0.470  &0.574\\ \hline
	16&Tuberculosis case detection rate (\%, all forms)&SH.TBS.DTEC.ZS&0.466  &0.609\\ \hline

			\end{tabular}
		}
	\end{centering}
\end{table*}

\end{document}